%% file: 0_main.tex
\documentclass{article}
\usepackage{style/unites}
\usepackage{XCharter}
\usepackage[scaled=1.1]{zlmtt} 
\usepackage{anyfontsize}

\input{style/macros}

\begin{document}

\makeatletter
\def\blfootnote{\gdef\@thefnmark{}\@footnotetext}
\makeatother

\makeatletter
\pagestyle{fancy}
\fancyhf{}
\renewcommand{\headrulewidth}{1pt}
\chead{\small\bf \input{1_title}
}
\cfoot{\thepage}
\thispagestyle{fancy}
\makeatother

\makeatletter
\def\icmldate#1{\gdef\@icmldate{#1}}
\icmldate{\today}
\makeatother

\makeatletter
\fancypagestyle{fancytitlepage}{
  \fancyhead{}
  \lhead{\includegraphics[height=0.8cm]{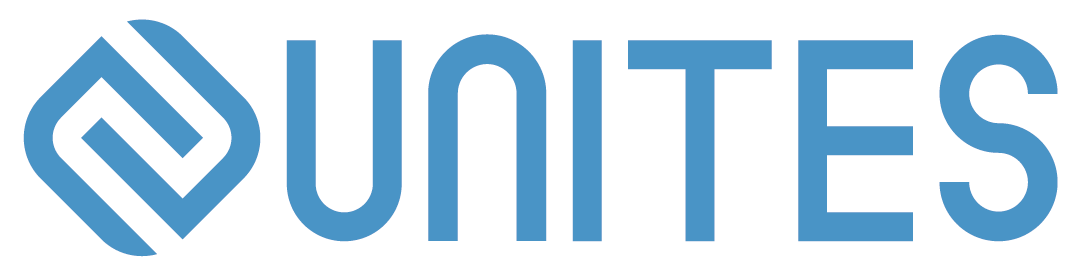}}
  \rhead{\it \@icmldate}
  \cfoot{}
}
\makeatother

\thispagestyle{fancytitlepage}

\vspace*{0.5em}

\noindent
\begin{titleblock}
    {\setlength{\parskip}{0cm}
     \raggedright
     {\setstretch{1.2}
      \LARGE\bfseries
      \input{1_title}
      \par}
    }
    \vskip 0.2cm
    
    \input{2_authors}
    \vskip 0.2cm
    
\begin{abstract}
Imputing physiological time series (arterial blood pressure, blood glucose, etc.) is essential for addressing the missingness that pervades clinical data. Yet modern imputation methods perform poorly in this domain: a recent benchmark found that simple linear interpolation outperformed \textit{every}
learned imputer on real-world clinical signals with realistic gaps.
We show that this reflects two properties of physiological missingness that generic
imputers ignore: gaps may occur when the signal is clinically extreme rather than typical, and gap lengths can easily span orders of magnitude. To this end, we introduce \textbf{Curriculum-Aware Interpolate-then-Refine} (\textbf{CAIR}), a two-stage
framework for physiological time-series imputation.
Our key motivation is to learn a coarse base curve and then repeatedly correct it toward physiological realism, rather than predict a gap in a single pass.
Consequently, CAIR\ couples a bidirectional-GRU interpolator with a
Transformer refiner that corrects its own estimate over three successive passes,
trained jointly under a broad, signal-agnostic random-gap curriculum.
We evaluate imputers stratified by gap length and missingness mechanism (MCAR, MAR, NMAR) rather than by a single average, and CAIR is the most accurate under every mechanism on continuous glucose monitoring (AI-READI) and arterial pressure in intensive care (MIMIC-III). Its margin over the strongest baseline grows with difficulty, from 9\% under MCAR to 19\% under value-dependent dropout, where generic learned imputers are weakest. We further show low reconstruction error alone does not recover the burden metrics clinicians act on: interpolants matching CAIR's error fail to preserve those metrics, imputers that recover them are far less accurate, and CAIR alone ranks among the best on both axes.
\end{abstract}
\end{titleblock}

\blfootnote{%
$^{\textrm{\Letter}}$ Corresponding authors: \{morris, haochenz, nick124, tianlong\}@cs.unc.edu
\\[2.5em]
\ifcsname @icmlpreprint\endcsname
  \textit{\csname @icmlpreprint\endcsname}%
\fi
}

\section{Introduction}
\label{sec:intro}
\input{1_intro}

\section{Related Work}
\label{sec:related}
\input{2_related}

\input{3_method}

\input{4_exp}

\input{5_conclusion}

\input{99_acknowledgement}

\newpage
\bibliography{999_ref}
\bibliographystyle{style/icml2025}

\titlespacing*{\section}{0pt}{*1}{*1}
\titlespacing*{\subsection}{0pt}{*1.25}{*1.25}
\titlespacing*{\subsubsection}{0pt}{*1.5}{*1.5}

\setlength{\abovedisplayskip}{\baselineskip} %
\setlength{\abovedisplayshortskip}{0.5\baselineskip} %
\setlength{\belowdisplayskip}{\baselineskip}
\setlength{\belowdisplayshortskip}{0.5\baselineskip}

\clearpage
\appendix
\label{sec:append}
\part*{Appendix}
{
\setlength{\parskip}{-0em}
\startcontents[sections]
\printcontents[sections]{ }{1}{}
}

\setlength{\parskip}{.5em}

\clearpage
\input{6_appendix}

\end{document}

%% file: style/macros.tex
\usepackage[utf8]{inputenc}
\usepackage[T1]{fontenc}
\usepackage{microtype}

\usepackage{amsmath}
\usepackage{amssymb}
\usepackage{amsfonts}
\usepackage{amsthm}
\usepackage{mathtools}
\usepackage{mathrsfs}
\usepackage{physics}
\usepackage{braket}
\usepackage{slashed}
\usepackage{nicefrac}
\usepackage{textcomp}
\usepackage{dsfont}
\usepackage{bbm}
\usepackage{bm}

\usepackage{graphicx}
\usepackage{subcaption}
\usepackage[export]{adjustbox}
\usepackage{float}
\usepackage{booktabs}
\usepackage{dcolumn}
\newcolumntype{d}[1]{D{.}{.}{#1}}
\usepackage{bigstrut, tabularx, multirow, makecell, diagbox}
\usepackage{colortbl}
\usepackage{tabularray}
\UseTblrLibrary{booktabs}
\usepackage{threeparttable}
\usepackage{tablefootnote}
\usepackage{fontawesome5}

\usepackage{placeins}
\usepackage{caption}
\usepackage{footnote}
\usepackage{enumitem}
\usepackage{multicol}
\usepackage{xspace}
\usepackage{titletoc}
\usepackage{titlesec}
\usepackage[bottom]{footmisc}
\usepackage{setspace}

\usepackage{wrapfig}
\usepackage{tikz}
\usepackage{quantikz}
\usepackage{dashbox}
\usepackage{mdframed}
\usepackage{marvosym}
\usepackage{pifont}
\usepackage{CJK}
\usepackage{url}

\usepackage[table,x11names]{xcolor}
\usepackage[most]{tcolorbox}
\tcbuselibrary{breakable}
\usetikzlibrary{decorations.pathreplacing, fit}

\definecolor{primaryblue}{HTML}{0066CC}
\definecolor{accentcyan}{HTML}{00D4AA}
\definecolor{warmorange}{HTML}{FF6B35}
\definecolor{deepgray}{HTML}{2C3E50}
\definecolor{lightgray}{HTML}{F8F9FA}
\definecolor{gradientstart}{HTML}{667eea}
\definecolor{gradientend}{HTML}{764ba2}

\definecolor{citecolor}{HTML}{0071bc}
\definecolor{citeblue}{RGB}{0, 113, 188}
\definecolor{linkcolor}{HTML}{9A4D92}
\definecolor{firebrick}{rgb}{0.698,0.133,0.133}

\definecolor{paleviolet}{HTML}{E1EEFC}
\definecolor{CarolinaUltraLight}{HTML}{E7F4FC}
\definecolor{lightgrey}{RGB}{247, 247, 247}
\definecolor{shadecolor}{HTML}{EFEFEF}
\definecolor{lightyellow}{rgb}{1.0, 0.95, 0.7}
\definecolor{lightblue}{rgb}{0.90, 0.95, 1.0}
\definecolor{light-gray}{gray}{0.95}

\definecolor{darkgrey}{rgb}{0.5, 0.5, 0.5}
\definecolor{darkgreen}{rgb}{0, 0.5, 0}
\definecolor{mydarkblue}{rgb}{0,0.08,0.45}
\definecolor{mydarkblue2}{rgb}{0.133, 0.133, 0.698}
\definecolor{echodrk}{HTML}{0099cc}
\definecolor{mymauve}{rgb}{0.58,0,0.82}
\definecolor{midnightblue}{rgb}{0.1,0.1,0.44}
\definecolor{oxfordblue}{rgb}{0.0,0.13,0.28}
\definecolor{prussianblue}{rgb}{0.0,0.19,0.33}
\definecolor{coolteal}{rgb}{0, 0.45, 0.45}
\definecolor{olive}{rgb}{0.1, 0.3, 0}
\definecolor{mypurple}{rgb}{0.5,0,0.5}
\definecolor{almond}{rgb}{0.94, 0.87, 0.8}

\definecolor{blue_ampEncoding}{HTML}{DAE8FC}
\definecolor{green_encoder}{HTML}{D5E8D4}
\definecolor{purple_decoder}{HTML}{E1D5E7}
\definecolor{yellow_measure}{HTML}{FFF2CC}
\definecolor{gray_block}{HTML}{F5F5F5}
\definecolor{pink_dru}{HTML}{FAD9D5}
\definecolor{orange_v}{HTML}{FAD7AC}

\definecolor{colorA}{rgb}{1,0,0}
\definecolor{colorB}{rgb}{0,0.3,1}
\definecolor{colorC}{rgb}{0.9,0.8,0.2}
\definecolor{colorD}{rgb}{0,0.65,0}
\definecolor{lesslightgray}{rgb}{0.5,0.5,0.5}
\definecolor{fundamental}{RGB}{55, 110, 111}
\definecolor{Gred}{RGB}{219, 50, 54}
\definecolor{ToCgreen}{RGB}{0, 128, 0}
\definecolor{Sepia}{RGB}{112, 66, 20}
\definecolor{Dblue}{rgb}{0,0.08,0.45}
\definecolor{Blue}{rgb}{0, 0, 0.8}
\definecolor{blue}{rgb}{0,0,1}
\definecolor{UNCblue!10}{rgb}{0.84,0.91,0.98}
\definecolor{RowAlt}{rgb}{0.98,0.98,0.99}

\definecolor{CarolinaBlue}{HTML}{7BAFD4}        % Official UNC Carolina Blue
\definecolor{CarolinaLightBlue}{HTML}{B3D4E5}   % Lighter version
\definecolor{CarolinaUltraLight}{HTML}{E8F4F8}  % Very light for background
\definecolor{CarolinaText}{HTML}{1C2B33}        % Dark text color

\usepackage[pagebackref=true,breaklinks=true,colorlinks,hyperfootnotes=false]{hyperref}
\hypersetup{
  colorlinks,
  citecolor=citeblue,
  linkcolor=firebrick,
  urlcolor=firebrick
}
\usepackage[nameinlink,capitalize,noabbrev]{cleveref}

\titlespacing\section{0pt}{4pt plus 4pt minus 2pt}{-2pt plus 2pt minus 2pt}
\titlespacing\subsection{0pt}{2pt plus 4pt minus 2pt}{-2pt plus 2pt minus 2pt}
\titlespacing\subsubsection{0pt}{2pt plus 4pt minus 2pt}{-2pt plus 2pt minus 2pt}

\makeatletter
\def\th@remark{%
  \thm@headfont{\bfseries}%
  \normalfont % body font
  \thm@preskip\topsep \divide\thm@preskip\tw@
  \thm@postskip\thm@preskip
}
\makeatother

\theoremstyle{definition}

\tcolorboxenvironment{theorem}{
  breakable,
  colback=black!10,
  colframe=white,
  width=\linewidth, 
  enlarge left by=0pt,
  enlarge right by=0pt,
  boxsep=5pt,
  boxrule=0pt,
  left=0pt,right=0pt,top=0pt,bottom=0pt,
  arc=8pt,
  before skip=\topsep,
  after skip=\topsep
}

\tcolorboxenvironment{lemma}{
  breakable,
  colback=black!10,
  colframe=white,
  width=\linewidth,
  enlarge left by=0pt,
  enlarge right by=0pt,
  boxsep=5pt,
  boxrule=0pt,
  left=0pt,right=0pt,top=0pt,bottom=0pt,
  arc=8pt,
  before skip=\topsep,
  after skip=\topsep
}

\tcolorboxenvironment{corollary}{
  breakable,
  colback=black!10,
  colframe=white,
  width=\linewidth,
  enlarge left by=0pt,
  enlarge right by=0pt,
  boxsep=5pt,
  boxrule=0pt,
  left=0pt,right=0pt,top=0pt,bottom=0pt,
  arc=8pt,
  before skip=\topsep,
  after skip=\topsep
}

\tcolorboxenvironment{proposition}{
  breakable,
  colback=black!10,
  colframe=white,
  width=\linewidth,
  enlarge left by=0pt,
  enlarge right by=0pt,
  boxsep=5pt,
  boxrule=0pt,
  left=0pt,right=0pt,top=0pt,bottom=0pt,
  arc=8pt,
  before skip=\topsep,
  after skip=\topsep
}

\tcolorboxenvironment{definition}{
  breakable,
  colback=black!10,
  colframe=white,
  width=\linewidth,
  enlarge left by=0pt,
  enlarge right by=0pt,
  boxsep=5pt,
  boxrule=0pt,
  left=0pt,right=0pt,top=0pt,bottom=0pt,
  arc=8pt,
  before skip=\topsep,
  after skip=\topsep
}

\tcolorboxenvironment{assumption}{
  breakable,
  colback=black!10,
  colframe=white,
  width=\linewidth,
  enlarge left by=0pt,
  enlarge right by=0pt,
  boxsep=5pt,
  boxrule=0pt,
  left=0pt,right=0pt,top=0pt,bottom=0pt,
  arc=8pt,
  before skip=\topsep,
  after skip=\topsep
}

\tcolorboxenvironment{claim}{
  breakable,
  colback=black!10,
  colframe=white,
  width=\linewidth,
  enlarge left by=0pt,
  enlarge right by=0pt,
  boxsep=5pt,
  boxrule=0pt,
  left=0pt,right=0pt,top=0pt,bottom=0pt,
  arc=8pt,
  before skip=\topsep,
  after skip=\topsep
}

\tcolorboxenvironment{problem}{
  breakable,
  colback=black!10,
  colframe=white,
  width=\linewidth,
  enlarge left by=0pt,
  enlarge right by=0pt,
  boxsep=5pt,
  boxrule=0pt,
  left=0pt,right=0pt,top=0pt,bottom=0pt,
  arc=8pt,
  before skip=\topsep,
  after skip=\topsep
}

\tcolorboxenvironment{question}{
  breakable,
  colback=black!10,
  colframe=white,
  width=\linewidth,
  enlarge left by=0pt,
  enlarge right by=0pt,
  boxsep=5pt,
  boxrule=0pt,
  left=0pt,right=0pt,top=0pt,bottom=0pt,
  arc=8pt,
  before skip=\topsep,
  after skip=\topsep
}

\newtcolorbox{titleblock}{
  enhanced,
  frame hidden,
  colback=CarolinaUltraLight,
  colframe=CarolinaUltraLight,
  boxrule=0pt,
  arc=10pt,
  left=14pt,
  right=14pt,
  top=14pt,
  bottom=14pt,
  width=\linewidth,
  before skip=12pt plus 4pt,
  after skip=12pt plus 4pt,
  grow to left by=1.5pt,
  grow to right by=1.5pt,
  before upper={
    \setlength{\parindent}{0cm}
    \setlength{\parskip}{0.5cm}
  }
}

\crefname{theorem}{Theorem}{Theorems}
\crefname{proposition}{Proposition}{Propositions}
\crefname{lemma}{Lemma}{Lemmas}
\crefname{corollary}{Corollary}{Corollaries}
\crefname{definition}{Definition}{Definitions}
\crefname{assumption}{Assumption}{Assumptions}
\crefname{remark}{Remark}{Remarks}
\crefname{problem}{Problem}{Problems}
\crefname{property}{Property}{property}
\crefname{question}{Question}{Questions}

\numberwithin{equation}{section}
\numberwithin{theorem}{section}
\numberwithin{proposition}{section}
\numberwithin{definition}{section}
\numberwithin{lemma}{section}
\numberwithin{assumption}{section}
\numberwithin{remark}{section}

\def\1{\bm{1}}

\makeatletter
\let\save@mathaccent\mathaccent
\newcommand*\if@single[3]{%
    \setbox0\hbox{${\mathaccent"0362{#1}}^H$}%
    \setbox2\hbox{${\mathaccent"0362{\kern0pt#1}}^H$}%
    \ifdim\ht0=\ht2 #3\else #2\fi
}
\newcommand*\rel@kern[1]{\kern#1\dimexpr\macc@kerna}
\newcommand*\widebar[1]{\@ifnextchar^{{\wide@bar{#1}{0}}}{\wide@bar{#1}{1}}}
\newcommand*\wide@bar[2]{\if@single{#1}{\wide@bar@{#1}{#2}{1}}{\wide@bar@{#1}{#2}{2}}}
\newcommand*\wide@bar@[3]{%
    \begingroup
    \def\mathaccent##1##2{%
        \let\mathaccent\save@mathaccent
        \if#32 \let\macc@nucleus\first@char \fi
        \setbox\z@\hbox{$\macc@style{\macc@nucleus}_{}$}%
        \setbox\tw@\hbox{$\macc@style{\macc@nucleus}{}_{}$}%
        \dimen@\wd\tw@
        \advance\dimen@-\wd\z@
        \divide\dimen@ 3
        \@tempdima\wd\tw@
        \advance\@tempdima-\scriptspace
        \divide\@tempdima 10
        \advance\dimen@-\@tempdima
        \ifdim\dimen@>\z@ \dimen@0pt\fi
        \rel@kern{0.6}\kern-\dimen@
        \if#31
        \overline{\rel@kern{-0.6}\kern\dimen@\macc@nucleus\rel@kern{0.4}\kern\dimen@}%
        \advance\dimen@0.4\dimexpr\macc@kerna
        \let\final@kern#2%
        \ifdim\dimen@<\z@ \let\final@kern1\fi
        \if\final@kern1 \kern-\dimen@\fi
        \else
        \overline{\rel@kern{-0.6}\kern\dimen@#1}%
        \fi
    }%
    \macc@depth\@ne
    \let\math@bgroup\@empty \let\math@egroup\macc@set@skewchar
    \mathsurround\z@ \frozen@everymath{\mathgroup\macc@group\relax}%
    \macc@set@skewchar\relax
    \let\mathaccentV\macc@nested@a
    \if#31
    \macc@nested@a\relax111{#1}%
    \else
    \def\gobble@till@marker##1\endmarker{}%
    \futurelet\first@char\gobble@till@marker#1\endmarker
    \ifcat\noexpand\first@char A\else
    \def\first@char{}%
    \fi
    \macc@nested@a\relax111{\first@char}%
    \fi
    \endgroup
    }
\makeatother

\DeclareMathAlphabet{\mathsfit}{\encodingdefault}{\sfdefault}{m}{sl}
\SetMathAlphabet{\mathsfit}{bold}{\encodingdefault}{\sfdefault}{bx}{n}

\let\hat\widehat

%% file: 1_title.tex
Curriculum-Aware Interpolate-then-Refine: Learned Physiological Time-Series Imputation under Realistic Missingness

%% file: 2_authors.tex
\begin{icmlauthorlist}
\mbox{Yu-Chao Huang},
Haochen Zhang, 
Nicholas Konz,
and
\mbox{Tianlong Chen}
\end{icmlauthorlist}

$^{1\,}$UNITES Lab, University of North Carolina at Chapel Hill  

%% file: 1_intro.tex
\input{fig1_overview_tikz.tex}
Physiological time series, such as continuous glucose monitoring (CGM), heart
rate, respiration and intensive-care vital signs, are the raw material of modern
data-driven medicine, and they are rarely complete
\citep{pratap_retention,braem_dataloss,ehr_dataquality}.
Sensors detach, wearables run out of battery, patients move, and clinical
devices are disconnected during procedures \citep{braem_dataloss,bent_ppg}.
Every downstream step, from computing a glycemic burden metric
\citep{kok_cgm_missing} to training a risk model \citep{hurst_imputation},
first has to decide what the missing values were.
Imputation is therefore not a preprocessing detail but a modeling choice that
propagates into every clinical conclusion drawn from the data
\citep{cgm_missing_metrics,icu_imputation_bench}.

The obvious remedy is a learned imputer, and a rich family now exists
\citep{brits,saits,csdi,timesnet}.
Yet a recent real-world benchmark reports a negative result: on clinical
signals with realistic gaps, \emph{linear interpolation} outperforms every method
tested, including deep learning models \citep{toye}.
We argue this is not evidence that learning cannot help, but a symptom of two
properties of physiological missingness that generic imputers and the standard
evaluation protocol both ignore:

\begin{enumerate}[leftmargin=2.3em]
\item [(C1)]
    \label{item:C1}
    \textbf{Missingness is mechanism-driven, not random.}
    We adopt the taxonomy of \citet{rubin}, under which missingness is
    \emph{missing completely at random} (MCAR) when the probability that a
    sample is lost is independent of the data, \emph{missing at random} (MAR)
    when that probability depends only on observed quantities such as a
    co-recorded covariate, and \emph{missing not at random} (NMAR) when it
    depends on the missing value itself.
    Physiological sensors fail more often during activity and during the
    clinical extremes that matter most, so their gaps are MAR or NMAR rather
    than MCAR.
    Imputers trained and scored under MCAR are therefore optimized for the one
    regime that never occurs.
\item [(C2)]
    \label{item:C2}
    \textbf{Gap length spans orders of magnitude.}
    A trace contains both single dropped samples and multi-hour holes.
    Averaging error over a fixed masking protocol lets the rare long gaps
    dominate the mean, so a single score can rank a method first while it is
    strictly worse in the short-gap regime required in clinical practice.
\end{enumerate}

We introduce \textbf{Curriculum-Aware Interpolate-then-Refine} (\textbf{CAIR}), a two-stage
neural imputer designed around these two properties.
Our key motivation is that the strength of classical interpolation is a
\emph{starting point}, not a ceiling: a coarse curve that respects the observed
endpoints is straightforward to construct, and the challenge is to correct it
toward physiological realism.
A deterministic interpolant, however, fixes that starting point in advance:
whatever it places inside a gap is a function of the two endpoints alone, which
makes it accurate over a few missing samples but unable to express a
multi-hour excursion.
CAIR\ therefore \emph{learns} the base curve rather than computing it.
Stage~1 is a bidirectional GRU \citep{gru,birnn} that predicts a base curve at
every
position from the observed values and the mask.
Stage~2 is a Transformer encoder \citep{transformer} that consumes this base
curve (both as conditioning and as a residual), together with any other
modalities the dataset provides, and corrects it over three successive
refinement passes, each conditioned on the previous estimate.

To address (C1), CAIR\ is trained under a signal-agnostic random-gap
curriculum that mixes scattered dropouts with contiguous blocks, rather than
under the single masking pattern used at evaluation time.
The training gap distribution proves as important as the architecture, which
the name of the method reflects: the \emph{same} network attains higher error
than linear interpolation when trained on masks tailored to the physiology of a
source domain, and lower error under all three mechanisms when trained on the
broad curriculum (Sec.~\ref{sec:exp-mimic}).
To address (C2), we evaluate every method stratified by gap length and by
missingness mechanism instead of reporting one average, and we show that the
ranking of methods depends on the length regime, so a single averaged score
cannot express it (Sec.~\ref{sec:exp-shortrange}).

\paragraph{Contributions.}
Our contributions are threefold:
\begin{itemize}[leftmargin=1.2em, itemsep=1pt, topsep=1pt]
    \item
    \textbf{Methodologically}, we propose CAIR, a two-stage imputer that
    replaces the deterministic interpolant with a jointly trained learned
    interpolator, and refines it with an iteratively unrolled Transformer.
    The design targets the two properties of physiological missingness above
    (C1, C2) rather than generic sequence modeling.

    \item
    \textbf{Empirically}, across two clinical domains, CGM from AI-READI
    \citep{aireadi} and arterial pressure from MIMIC-III \citep{mimic3},
    CAIR\ attains the lowest reconstruction error of every method we evaluate
    under all three mechanisms, and its margin over the strongest baseline
    \emph{grows} with difficulty: $9\%$ under MCAR, $16\%$ under MAR, and $19\%$
    under NMAR.
    The generic learned imputers that \citet{toye} report as failures fail here
    too, which isolates how the model is designed and trained, rather than
    neural capacity, as the cause of the gain.

    \item
    \textbf{Analytically}, we show that low reconstruction error and faithful
    downstream clinical metrics are distinct axes.
    Interpolants that match the reconstruction error of CAIR\ fail to preserve
    the burden metrics clinicians read (linear interpolation recovers $0.14$ of
    the recoverable burden on MIMIC-III), while tabular and neural imputers that
    do recover the burden incur $10$--$60\%$ higher error.
    CAIR\ is the only method that ranks among the best on both axes.
\end{itemize}

%% file: 2_related.tex
\paragraph{Classical and statistical imputation.}
Deployed clinical pipelines still rely on deterministic interpolants: linear
fills, shape-preserving cubics such as PCHIP \citep{pchip} and akima
\citep{akima}, and smoothers such as Savitzky--Golay \citep{savgol}.
They carry no fitting cost and remain accurate on short gaps, which is why they
are still the default for CGM metric computation \citep{cgm_missing_metrics}.
Statistical imputers (MICE \citep{mice}, missForest \citep{missforest},
$k$NN and hot-deck) instead treat the series as a table of features, recovering
distributional structure that interpolation discards at the cost of temporal
smoothness.
Both families share the property CAIR\ targets: what they place inside a gap is
determined in advance, either by the two endpoints or by a marginal
distribution, and cannot be adapted to the model that consumes it.
CAIR\ keeps the interpolate-then-refine structure that motivates the classical
prior, but makes the first stage learned and trainable end-to-end, so the base
curve adapts to the refiner.

\paragraph{Learned time-series imputation.}
Recurrent imputers exploit informative missingness directly: GRU-D uses decay
toward the empirical mean \citep{gru_d}, BRITS imputes bidirectionally with
consistency between directions \citep{brits}, and M-RNN combines intra- and
inter-stream recurrence \citep{mrnn}.
Attention-based and generative approaches followed.
Closest to CAIR, SAITS \citep{saits} estimates with one diagonally-masked
self-attention block, writes those estimates back into the input, and
re-estimates with a second, supervising both; we retain that structure but make
the first stage an information-restricted recurrent interpolator, unroll a single
tied-weight refiner over it, and mark the model's own fills with a provenance
code at every pass.
A separate line makes the interpolation step itself learned: interpolation
prediction networks \citep{ipn} and mTAN \citep{mtan} attach a learned
interpolation layer to a downstream network and train the pair end-to-end.
Those layers are kernel smoothers designed to place irregular samples on a
regular grid, and they feed a classifier; CAIR's Stage~1 is a recurrent
sequence model supervised directly against held-out values, and what it feeds is
a refiner that corrects it.
GP-VAE \citep{gpvae} places a Gaussian-process prior in a VAE latent space,
CSDI \citep{csdi} runs conditional score-based diffusion, and backbones such as
TimesNet \citep{timesnet} treat imputation as one of several tasks.
All are domain-agnostic, and our results suggest physiological imputation needs
a domain-specific model: M-RNN and GP-VAE fall below every non-constant
baseline on both CGM and ICU vitals.

\paragraph{Evaluating under realistic missingness.}
Rubin's MCAR/MAR/NMAR taxonomy \citep{rubin} is standard in statistics but is
rarely used to structure machine-learning imputation benchmarks, which typically
delete values completely at random.
Recent work pushes back: \citet{beyond_random_missingness} and
\citet{icu_imputation_bench} both show that clinically plausible missingness
patterns change method rankings.
Closest to our work, \citet{toye} evaluate eleven imputers on real-world clinical
signals under mechanism-driven deletion and find that linear interpolation
outperforms all of them.
We adopt their protocol verbatim as our hardest evaluation, and show that their
result is a statement about generic imputers rather than about learned
imputation in general.
Detailed discussion is in the supplementary material.

%% file: 3_method.tex
\section{Method}
\label{sec:method}

\subsection{Problem Setup}
A physiological trace is a uniformly sampled series $x \in \mathbb{R}^{T}$ with
observation mask $m \in \{0,1\}^{T}$ ($m_t{=}1$ if $x_t$ is observed; 5-minute
grid throughout).
We write $\odot$ for the elementwise product, $e \in \{0,1\}^{T}$ for the
held-out evaluation mask, and $\hat{x}$ for the imputed trace; a \emph{gap} is a
maximal run of consecutive missing positions and its \emph{length} is that run's
size.
Imputation predicts $\hat{x}_t$ at every missing position ($m_t{=}0$) from the
observed ones.
Values are $z$-normalized with training statistics and rescaled to native units
for reporting.
When auxiliary channels are available (co-recorded vitals, activity, sleep
state), they are appended to the per-position conditioning and the model is
otherwise unchanged.

\subsection{Interpolate-then-Refine Architecture}
CAIR\ has two neural stages, illustrated in Fig.~\ref{fig:overview}d.

\paragraph{Stage 1: Learned interpolator.}
A bidirectional GRU \citep{gru,birnn} reads the masked signal together with
its mask,
$f_t = [\,x_t m_t,\; m_t\,]$, and predicts a base value at \emph{every}
position,
\begin{equation}
  y^{0} = \mathrm{Head}\big(\mathrm{BiGRU}(f)\big) \in \mathbb{R}^{T},
\end{equation}
with hidden size $128$ and $4$ layers, where $\mathrm{Head}$ is a single linear
map from the $256$-dimensional concatenated forward and backward hidden state to
one value per position.
This is the component that replaces the deterministic interpolant: where a
classical pipeline computes a PCHIP or akima curve from the two gap endpoints,
$y^{0}$ is trained.
The output head is zero-initialized, so at the start of training the model
reduces to its refiner and learns the base curve as a residual correction.

\paragraph{Stage 2: Transformer refiner.}
A bidirectional pre-norm Transformer encoder \citep{transformer}
($d_{\text{model}}{=}128$,
$8$ layers, $8$ heads, FFN width $512$) corrects the base curve.
The interpolator output enters twice: once inside the per-position conditioning,
occupying the slots a classical pipeline reserves for its interpolant, and once
as an additive residual base,
\begin{equation}
  \hat{x} = y^{0} + \mathrm{Transformer}\big(\phi(x\!\odot\!m,\, m,\, y^{0},\, \tau)\big),
\end{equation}
where $\phi$ is the per-position conditioning built from the observed samples,
the mask, learned time-of-day embeddings $\tau$, and $y^{0}$.
The two stages are assigned complementary roles: the GRU produces a smooth base
curve, and the Transformer adds the data-driven physiological shape
(post-prandial rises, nocturnal dips, pressure excursions) that a smooth base
cannot express.

\subsection{Training Objective}
\label{sec:loss}
CAIR\ is trained from scratch under a \emph{signal-agnostic random-gap
curriculum}: at each step a mixture of scattered points and contiguous blocks
($\sim$$20\%$ of observed samples) is held out and reconstructed.
The curriculum is deliberately not specialized for the typical failure modes of
any single signal type, which keeps the method general (C1).
Indeed, Sec.~\ref{sec:exp-mimic} shows that using masks tailored to the
physiology of a \emph{source} domain results in worse generalization to new
signals.

The refiner is unrolled for three passes, each conditioned on the previous
prediction, under an increasing weight schedule that emphasizes the final pass.
An auxiliary term supervises the Stage-1 output directly so it learns a smooth
base rather than collapsing into the refiner:
\begin{equation}
  \mathcal{L} = \sum_{k=1}^{3} w_k \big\lVert (\hat{x}^{(k)} - x)\odot e \big\rVert^2
  + \lambda_{\text{aux}} \big\lVert (y^{0} - x)\odot e \big\rVert^2 ,
\end{equation}
with $w = (0.15,\,0.35,\,0.50)$ and $\lambda_{\text{aux}}{=}0.7$.
Optimization uses AdamW \citep{adamw} at learning rate $3\times10^{-4}$ with
cosine decay, and we keep an exponential moving average of the weights
(decay $0.999$) for inference \citep{ema_swa}.

\subsection{Inference}
At test time CAIR\ slides a $576$-step window over the full trace
(stride $144$) with cosine-window averaging and the same three refiner passes
used in training (a base pass plus two conditioned on the previous estimate);
predictions at observed positions are left unchanged.

%% file: 4_exp.tex
\IfFileExists{ablation_numbers.tex}{\input{ablation_numbers}}{}
\providecommand{\TBD}{\textcolor{red}{\,--.--\,}}
\providecommand{\aNpart}{\textcolor{red}{NN}}
\providecommand{\aFloor}{0.16}
\providecommand{\aFullMcar}{\TBD}\providecommand{\aFullMar}{\TBD}
\providecommand{\aFullNmar}{\TBD}\providecommand{\aFullAll}{\TBD}
\providecommand{\aAuxMcar}{\TBD}\providecommand{\aAuxMar}{\TBD}
\providecommand{\aAuxNmar}{\TBD}\providecommand{\aAuxAll}{\TBD}
\providecommand{\aResMcar}{\TBD}\providecommand{\aResMar}{\TBD}
\providecommand{\aResNmar}{\TBD}\providecommand{\aResAll}{\TBD}
\providecommand{\aAtnMcar}{\TBD}\providecommand{\aAtnMar}{\TBD}
\providecommand{\aAtnNmar}{\TBD}\providecommand{\aAtnAll}{\TBD}
\providecommand{\aLinMcar}{\TBD}\providecommand{\aLinMar}{\TBD}
\providecommand{\aLinNmar}{\TBD}\providecommand{\aLinAll}{\TBD}
\providecommand{\aDAux}{\TBD}\providecommand{\aDRes}{\TBD}
\providecommand{\aDAtn}{\TBD}\providecommand{\aDLin}{\TBD}

\section{Experiments}
\label{sec:exp}

We ask five questions in turn.
Does CAIR\ outperform linear interpolation under \emph{realistic} missingness
(Sec.~\ref{sec:exp-mech})?
Does the answer depend on gap length, and is that dependence visible in a single
averaged score (Sec.~\ref{sec:exp-shortrange}--\ref{sec:exp-artifact})?
Does the advantage survive a change of signal and clinical domain
(Sec.~\ref{sec:exp-other}--\ref{sec:exp-mimic})?
Does lower reconstruction error recover the metrics clinicians read
(Sec.~\ref{sec:exp-down})?
And which components of the model are responsible
(Sec.~\ref{sec:exp-ablation})?
Full preprocessing, hyperparameters and protocol details are in the supplementary material.

\subsection{Setup}
\label{sec:exp-setup}

\paragraph{Data.}
\textbf{AI-READI} \citep{aireadi,aireadi_dataset} provides CGM, heart rate and
respiration for $2{,}280$ participants, split $1{,}576$ train / $352$ validation
/ $352$ test.
The imputation target is day two of each trace, a $24$\,h window on the 5-min
grid.
\textbf{MIMIC-III} \citep{mimic3} provides intensive-care vitals; we assemble
$22{,}156$ twenty-four-hour windows on the same grid with the observed sensor
masks, split subject-disjoint, and impute arterial blood pressure (ABP) and
heart rate (HR).

\paragraph{Missingness protocols.}
Following \citet{toye} we simulate three mechanisms \citep{rubin} on the target
window, each at six rates from $5\%$ to $30\%$: \emph{MCAR} (independent
deletion), \emph{MAR} (contiguous windows triggered by a co-recorded covariate
(time-aligned wearable activity on AI-READI, a covariate vital on MIMIC-III),
and \emph{NMAR} (windows triggered by clinically extreme values of the target
itself: $<70$ or $>150$\,mg/dL for glucose).
The \emph{gap-length protocol} carves a single contiguous gap of fixed length
$L\in\{3,6,9,12\}$ samples ($15$--$60$\,min) at observed positions,
$10$ placements per participant and length, and scores only the held-out points.
The \emph{physiological protocol} masks $20\%$ of samples in event-aligned
blocks under five strategies (meal-post, sleep, ascending, dipping, combined).

\paragraph{Baselines.}
We compare against twenty baselines in four families: constant and
interpolation fills (linear, LOCF \citep{locf}, mean, mode); shape-preserving
and smoothing methods (PCHIP \citep{pchip}, akima \citep{akima}, cubic spline,
Savitzky--Golay \citep{savgol}, EWMA, a Kalman smoother \citep{kalman},
truncated-Fourier reconstruction, bidirectional AR); tabular imputers (MICE
\citep{mice}, missForest \citep{missforest}, $k$NN \citep{knn_impute}, hot-deck
\citep{hotdeck}); and learned sequence imputers (SAITS \citep{saits}, BRITS
\citep{brits}, M-RNN \citep{mrnn}, GP-VAE \citep{gpvae}), the last group
retrained on the same windows as CAIR. All see identical masks and seeds;
per-protocol subsets and settings are in the supplement.

\paragraph{Metrics.}
We report RMSE in native units (mg/dL, mmHg, bpm, breaths/min).
For downstream quality we report the \emph{metric-recovery ratio} (MRR): one
minus a method's error on a clinical metric divided by the mean-fill error on
that metric, so $1$ is perfect recovery, $0$ is no better than mean imputation,
and negative values are worse. We restrict MRR to the shape and variability
metrics that mean imputation fails to preserve (time in, above and below range
\citep{battelino_tir}, MAGE \citep{mage}, coefficient of variation); on
mean-preserving metrics the denominator degenerates.

\subsection{Main Results}
\label{sec:exp-mech}

Table~\ref{tab:mech} is the main result.
Under all three mechanisms CAIR\ attains the lowest RMSE, and its margin over
linear interpolation, the method \citet{toye} found strongest, \emph{grows} with
difficulty: $9\%$ under MCAR, $16\%$ under MAR, and $19\%$ under NMAR, the
hardest and most clinically loaded regime.

One of the most striking results is the contrast with the neural baselines.
M-RNN and GP-VAE fall below every non-constant baseline under all three
mechanisms, reproducing the failure \citet{toye} report.
Neural capacity therefore cannot be what separates CAIR\ from them; the
difference is that CAIR\ is trained under a gap distribution broad enough to
cover the failure modes that occur in practice (C1).
Sec.~\ref{sec:exp-mimic} tests this claim directly by retraining the identical
network on masks designed for a different signal's physiology.

\begin{table}[t]
  \centering
  {\small
  \setlength{\tabcolsep}{5pt}
  \begin{tabular}{lccc}
    \toprule
    Method & MCAR $\downarrow$ & MAR $\downarrow$ & NMAR $\downarrow$ \\
    \midrule
    \textbf{CAIR\ (Ours)} & \textbf{2.66} & \textbf{10.33} & \textbf{23.50} \\
    \midrule
    linear interp & \underline{2.91} & \underline{12.34} & \underline{28.94} \\
    MICE          & 3.26 & 19.83 & 49.35 \\
    missForest    & 3.33 & 16.54 & 40.90 \\
    hot-deck      & 5.04 & 16.06 & 43.44 \\
    $k$NN         & 5.19 & 15.45 & 43.54 \\
    LOCF          & 6.22 & 19.84 & 34.44 \\
    Fourier       & 9.91 & 18.47 & 32.72 \\
    mean          & 28.59 & 28.81 & 54.86 \\
    GP-VAE        & 26.46 & 40.49 & 70.39 \\
    M-RNN         & 44.39 & 44.34 & 64.63 \\
    \bottomrule
  \end{tabular}}
  \caption{\textbf{Realistic missingness imputation on AI-READI CGM} (all $352$ test
  participants; RMSE mg/dL averaged over six missingness rates, lower is
  better). CAIR\ is best under all three mechanisms and its margin grows with
  difficulty. The two generic learned imputers fall below every non-constant
  baseline throughout. \textbf{Bold} = best, \underline{underline} = second best, per column.}
  \label{tab:mech}
\end{table}

\subsection{Gap Length}
\label{sec:exp-shortrange}

Imputation beyond roughly one hour is not a realistic clinical target: once a
gap spans an excursion, the in-gap information is not present in the endpoints.
Standard CGM pipelines accordingly interpolate only gaps shorter than
$30$--$45$\,min and segment the trace beyond that \citep{glucobench}.
We therefore re-score every method on single contiguous gaps of $15$--$60$\,min
(Table~\ref{tab:shortgap}).

CAIR\ is superior to the other methods on average across these gap lengths,
being best at the longer intervals ($45$ and $60$\,min) and a very close
runner-up at the shorter ones ($15$ and $30$\,min).

\begin{table}[t]
  \centering
  {\small
  \setlength{\tabcolsep}{3.6pt}
  \begin{tabular}{lccccc}
    \toprule
    Method & 15\,min & 30\,min & 45\,min & 60\,min & \textbf{mean} $\downarrow$ \\
    \midrule
    \textbf{CAIR\ (Ours)}   & \underline{2.84} & \underline{5.03} & \textbf{6.66} & \textbf{8.23} & \textbf{5.69} \\
    akima                    & \textbf{2.76} & \textbf{5.01} & \underline{6.87} & 8.61 & \underline{5.81} \\
    PCHIP                    & 2.84 & 5.14 & 6.92 & \underline{8.58} & 5.87 \\
    linear                   & 3.20 & 5.64 & 7.45 & 9.09 & 6.34 \\
    SAITS                    & 4.60 & 7.54 & 9.76 & 11.79 & 8.42 \\
    BRITS                    & 6.25 & 9.42 & 11.67 & 13.33 & 10.17 \\
    \bottomrule
  \end{tabular}}
  \caption{\textbf{Imputation performance vs. gap length} on AI-READI CGM (all $352$ test
  participants; RMSE mg/dL). CAIR\ is best on the mean and at
  $45$--$60$\,min, and second by a small margin at $15$--$30$\,min.
  \textbf{Bold} = best, \underline{underline} = second, per column.}
  \label{tab:shortgap}
\end{table}

\begin{figure*}[t]
  \centering
  \includegraphics[width=\textwidth]{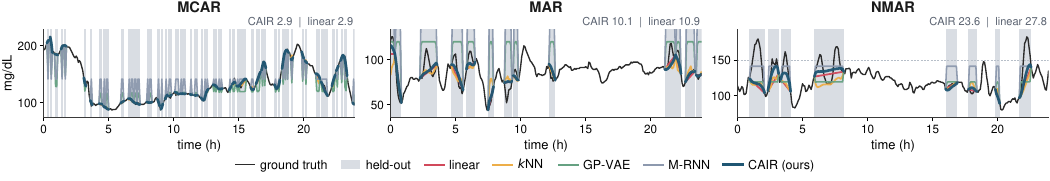}
  \caption{\textbf{Missingness mechanism examples.} (AI-READI CGM at $30\%$
  missing; five representative methods, drawn at held-out positions only). Each panel
  is its mechanism's median-difficulty window; insets give in-gap RMSE (mg/dL).
  MCAR gaps are near-trivial; NMAR puts contiguous gaps on the excursions
  crossing the dashed $150$\,mg/dL line.}
  \label{fig:gallery}
\end{figure*}

\subsection{Physiological Masking}
\label{sec:exp-artifact}

Table~\ref{tab:phys} reports the physiological five-strategy protocol over the
full baseline suite.
CAIR\ attains the best average, improving on the strongest classical baseline
by $1.33$\,mg/dL, and is the most accurate method on the \emph{meal},
\emph{sleep} and \emph{combined} strategies.
The gain concentrates where a data-driven prior helps most: the long
\emph{sleep} blocks, which cost $24.9$\,mg/dL for CAIR\ against $29.3$ for the
strongest classical baseline.

\begin{table}[t]
  \centering
  {\small
  \setlength{\tabcolsep}{3.2pt}
  \begin{tabular}{lcccccc}
    \toprule
    Method & meal & sleep & asc & dip & comb & \textbf{AVG} $\downarrow$ \\
    \midrule
    \textbf{CAIR\ (Ours)} & \textbf{15.25} & \textbf{24.94} & 9.01 & 7.49 & \textbf{16.37} & \textbf{14.61} \\
    \midrule
    PCHIP           & \underline{16.56} & \underline{29.26} & \textbf{7.86} & 7.36 & 18.65 & \underline{15.94} \\
    linear          & 16.97 & 29.27 & 8.42 & \underline{7.31} & \underline{18.30} & 16.05 \\
    akima           & 17.82 & 36.73 & \underline{8.02} & \textbf{7.24} & 19.64 & 17.89 \\
    Kalman          & 17.97 & 34.70 & 8.88 & 8.09 & 20.29 & 17.99 \\
    Savitzky--Golay & 17.27 & 29.45 & 13.41 & 11.22 & 19.43 & 18.15 \\
    SAITS           & 21.27 & 32.02 & 11.22 & 10.22 & 19.70 & 18.89 \\
    BRITS           & 32.72 & 43.61 & 24.25 & 20.81 & 31.45 & 30.57 \\
    \bottomrule
  \end{tabular}}
  \caption{\textbf{Imputation of physiologically-significant missingness} on AI-READI (all $352$
  test participants; RMSE mg/dL, lower is better). CAIR\ attains the best
  average over the full baseline suite, and the gain concentrates in the long
  \emph{sleep} blocks. The four weakest classical fills are in the supplement.
  \textbf{Bold} = best, \underline{underline} = second
  best, per column.}
  \label{tab:phys}
\end{table}

\begin{table}[t]
  \centering
  {\small
  \setlength{\tabcolsep}{4pt}
  \begin{tabular}{@{}lccc@{}}
    \toprule
    \multicolumn{4}{@{}l}{\emph{(a) Physiological vs.\ random-gap training masks}}\\
    Training masks & \multicolumn{3}{c@{}}{ABP RMSE $\downarrow$}\\
    & MCAR & MAR & NMAR \\
    \midrule
    linear                  & \underline{4.40} & 9.23 & \underline{11.85} \\
    CAIR, CGM masks        & 5.47 & \underline{8.74} & 12.10 \\
    \textbf{CAIR, random-gap} & \textbf{4.30} & \textbf{8.60} & \textbf{11.42} \\
    \midrule
    \multicolumn{4}{@{}l}{\emph{(b) Clinical-burden recovery}}\\
    Method & \multicolumn{3}{c@{}}{MRR $\uparrow$}\\
    & MCAR & MAR & NMAR \\
    \midrule
    linear                & $-0.31$ & $-0.30$ & $+0.14$ \\
    $k$NN                 & $\mathbf{+0.79}$ & $\mathbf{+0.74}$ & $\mathbf{+0.65}$ \\
    \textbf{CAIR (ours)} & \underline{$+0.71$} & \underline{$+0.64$} & \underline{$+0.64$} \\
    \bottomrule
  \end{tabular}}
  \caption{\textbf{Two analyses behind the MIMIC-III results} (ABP).
  \textbf{(a)} The \emph{identical} architecture, trained on the same ABP data,
  is less accurate than linear under MCAR and NMAR when trained on masks carried
  over from glucose physiology, and more accurate under all three under the
  random-gap curriculum.
  \textbf{(b)} Linear is \emph{worse than mean-fill} ($\mathrm{MRR}<0$) under
  MCAR/MAR; CAIR\ and $k$NN recover most of the burden.
  \textbf{Bold} = best, \underline{underline} = second, per column.
  A third analysis, in which no imputer separates from any other when
  predicting mortality from arterial pressure alone, is reported in
  Sec.~\ref{sec:exp-down}.}
  \label{tab:analysis}
\end{table}

\subsection{Heart Rate and Respiration}
\label{sec:exp-other}

CAIR\ is not bespoke to any one type of time series such as blood
glucose, and is designed to be general.
Accordingly, in this section and in Sec.~\ref{sec:exp-mimic} we evaluate it in
two further physiological domains.
We apply it unchanged to the heart-rate and respiration channels of AI-READI
under the gap-length protocol, in two variants: \emph{univariate}, seeing only
the target channel, and \emph{multivariate}, also conditioned on the co-recorded
context the dataset provides (steps, energy expenditure, sleep state, stress,
and the complementary cardiorespiratory channel).

The two variants separate sharply (Table~\ref{tab:hr-resp}).
Univariate CAIR\ is \emph{worse than linear} at every gap length on both
signals; the multivariate variant reverses this, attaining the lowest RMSE
everywhere and improving on the strongest classical baseline by $20$--$34\%$ on
heart rate and $15$--$28\%$ on respiration.
By contrast, the \emph{same} channels leave CGM accuracy unchanged (a nine-rung
ladder moves its own average only $13.11\!\to\!12.98$\,mg/dL, inside seed noise;
supplement).
Glucose is autocorrelated and endogenously driven, so its own history already
carries what a covariate could add; heart rate and respiration are driven by
exogenous activity, so the covariate adds what the target's history lacks.
This yields a general rule: \emph{cross-modal conditioning helps exactly when the
target is exogenously driven}, and requires no change to the architecture.

\begin{table}[t]
\centering
  \setlength{\tabcolsep}{1mm}
  {\small
\begin{tabular}{lcccccccc}
\toprule
& \multicolumn{4}{c}{Heart rate (bpm) $\downarrow$} & \multicolumn{4}{c}{Resp.\ (br/min) $\downarrow$} \\
\cmidrule(lr){2-5}\cmidrule(lr){6-9}
Method & 15 & 30 & 45 & 60 & 15 & 30 & 45 & 60 \\
\midrule
Akima              & 4.54 & 5.93 & 6.86 & 7.87 & 2.73 & 3.23 & 3.72 & 4.30 \\
AR (bidir.)        & 4.71 & 5.62 & \underline{6.09} & \underline{6.72} & 2.77 & 3.09 & \underline{3.33} & \underline{3.57} \\
PCHIP              & 4.45 & 5.62 & 6.27 & 7.05 & 2.63 & 3.07 & 3.43 & 3.76 \\
Linear             & \underline{4.40} & \underline{5.52} & 6.12 & 6.86 & \underline{2.61} & \underline{3.00} & 3.34 & 3.65 \\
\midrule
\multicolumn{9}{l}{CAIR\ (ours)} \\
\quad univariate   & 4.96 & 6.49 & 6.74 & 7.51 & 2.90 & 3.92 & 4.50 & 5.21 \\
\quad \textbf{multivariate} & \textbf{3.50} & \textbf{4.19} & \textbf{4.29} & \textbf{4.43} & \textbf{1.88} & \textbf{2.54} & \textbf{2.62} & \textbf{2.72} \\
\bottomrule
\end{tabular}}
  \caption{\textbf{Cross-modal conditioning is what transfers} (AI-READI,
gap-length protocol, RMSE in native units, lower is better). Column headings are
gap lengths in minutes; ``Resp.'' is respiration and ``AR (bidir.)'' is
bidirectional AR. Univariate
CAIR\ is less accurate than linear at every gap length on both signals; adding
the co-recorded context makes it best everywhere. On glucose the same context
adds nothing (Sec.~\ref{sec:exp-other}). \textbf{Bold} = best, \underline{underline} = second, per column.}
  \label{tab:hr-resp}
\end{table}

\subsection{MIMIC-III}
\label{sec:exp-mimic}

AI-READI provides a single signal from a single sensor class.
We therefore repeat the three-mechanism protocol on MIMIC-III intensive-care
vitals, adapting the triggers to ICU physiology (MAR from a co-recorded
covariate vital, NMAR from clinically extreme target values).
These signals are smoother and more locally linear than glucose, which sets a
demanding standard for linear interpolation.

However, on arterial pressure CAIR\ is the most accurate of every method we evaluate under
every mechanism (Table~\ref{tab:mimic}); a paired Wilcoxon signed-rank test
against linear interpolation is significant in its favor under all three
($p<3\times10^{-3}$).
Heart rate is a boundary case: CAIR\ attains the lowest mean RMSE under MAR and
linear interpolation the lowest under MCAR and NMAR, but only the MCAR
difference is statistically significant, the MAR and NMAR differences being ties
($p=0.83$ and $p=0.37$).
This is consistent with a smooth signal on which interpolation is already
near-optimal.
Per-mechanism tests for both signals, including the heart-rate case where linear
interpolation is significantly ahead, are reported in the supplement.
As on CGM, M-RNN and GP-VAE fall far below every interpolant on both signals.

\paragraph{The training gap distribution is what drives the result.}
Both rows are trained on the same ABP data with the same architecture and
budget; only the training-time masking differs.
Under the hand-designed glucose-physiological masks the \emph{identical} model
is \emph{less accurate} than linear (MCAR $5.47$, NMAR $12.10$ mmHg); under the
signal-agnostic random-gap curriculum of Sec.~\ref{sec:loss}, specialized to no
signal in particular, it is more accurate under all three
(Table~\ref{tab:analysis}a).
The masks encoding one domain's failure modes are thus what does \emph{not}
carry over, and a deliberately broad gap distribution is what lets the
architecture transfer.

\begin{table}[t]
  \centering
  {\small
  \setlength{\tabcolsep}{1mm}
  \begin{tabular}{lcccccc}
    \toprule
    & \multicolumn{3}{c}{ABP-mean (mmHg) $\downarrow$} & \multicolumn{3}{c}{HR (bpm) $\downarrow$} \\
    \cmidrule(lr){2-4}\cmidrule(lr){5-7}
    Method & MCAR & MAR & NMAR & MCAR & MAR & NMAR \\
    \midrule
    \textbf{CAIR\ (Ours)} & \textbf{4.30} & \textbf{8.60} & \textbf{11.42} & \underline{2.26} & \textbf{5.28} & \underline{6.48} \\
    linear interp & \underline{4.40} & \underline{9.23} & \underline{11.85} & \textbf{2.22} & \underline{5.50} & \textbf{6.42} \\
    GP-VAE & 11.03 & 12.82 & 17.94 & 9.17 & 14.01 & 18.00 \\
    M-RNN & 13.45 & 13.19 & 18.35 & 15.28 & 14.53 & 18.60 \\
    \bottomrule
  \end{tabular}}
  \caption{\textbf{Cross-domain transfer to MIMIC-III ICU vitals}
  (reconstruction RMSE, mean over six missingness rates, lower is better).
  CAIR\ is the most accurate method on arterial pressure under every mechanism;
  heart rate is a boundary case. The CAIR\ row is the multivariate variant;
  both variants, the remaining baselines and per-mechanism Wilcoxon tests are
  in the supplement.
  \textbf{Bold} = best, \underline{underline} = second, per column.}
  \label{tab:mimic}
\end{table}

\subsection{Downstream Clinical Metrics}
\label{sec:exp-down}

Low reconstruction error does not by itself recover the metrics clinicians act
on.
We score clinical-burden recovery as MRR on the threshold metrics read at the
bedside (time in the normal band and time above and below it) under the
hardest NMAR mechanism.

These two axes split the methods into two groups, and CAIR\ is the only
method that ranks among the best on both (full table in the supplement).
Interpolants that match its RMSE \emph{fail to preserve} the burden: linear
recovers only $0.14$ of it and under MCAR is \emph{worse than mean-fill}
($-0.31$, Table~\ref{tab:analysis}b), smoothing away the excursions the
threshold metrics count.
Conversely, the tabular and neural imputers that match CAIR's burden recovery
($k$NN at $0.65$, then MICE, M-RNN, GP-VAE) carry $10$--$60\%$ higher RMSE, so
only CAIR\ attains both the lowest RMSE ($11.4$\,mmHg) and burden recovery among
the best ($0.64$).
The same ordering holds on CGM (time-in-range recovery $0.44$ vs.\ $0.36$ for
linear under NMAR \citep{battelino_tir,mage}) and in all four diabetes study
groups (supplement), which rules out differences in cohort composition as the
explanation.

\paragraph{Hard outcomes are insensitive to reconstruction fidelity.}
On stay-level outcomes the effect is mediated by the target vital, so we isolate
it with a single-vital classifier. Predicting mortality from arterial pressure
alone, any imputer improves on mean-fill and tracks the oracle ceiling
(supplement), but the fills do not separate: with a
shape-sensitive 1D-CNN readout every fill matches or \emph{exceeds} the oracle
trace (linear $0.692$ vs.\ oracle $0.679$ at $30\%$ MCAR), because interpolation
denoises the vital and a weak-signal classifier rewards smoothing.
The fidelity CAIR\ optimizes is thus the opposite of what single-vital outcome
prediction rewards, which argues that physiological imputation should be judged
by reconstruction and burden recovery, not hard-outcome AUROC.

\subsection{Qualitative Results}
\label{sec:exp-qual}

Aggregate error does not show \emph{how} methods fail; Fig.~\ref{fig:gallery}
does (both rates, MIMIC-III and a difficulty sweep are in the supplement).
Scattered masking is near-trivial ($1$--$4$\,mg/dL), long contiguous blocks far
harder ($30$--$100$\,mg/dL): once a block spans an excursion, every
method reverts to a near-flat fill.
The methods differ in \emph{what} they revert to: GP-VAE and M-RNN to a
training-set constant, linear to the chord its endpoints imply, and CAIR\ toward
the excursion without inventing one.
CAIR\ therefore degrades gracefully rather than hallucinating, which is the
visual basis for restricting clinical claims to short gaps.

\input{4b_ablation}

%% file: ablation_numbers.tex
\providecommand{\TBD}{\textcolor{red}{\,--.--\,}}

\providecommand{\aNpart}{352}       %
\providecommand{\aFloor}{0.16}

\providecommand{\aFullMcar}{2.66}\providecommand{\aFullMar}{10.33}
\providecommand{\aFullNmar}{23.50}\providecommand{\aFullAll}{12.16}

\providecommand{\aAuxMcar}{6.36}\providecommand{\aAuxMar}{10.75}
\providecommand{\aAuxNmar}{27.59}\providecommand{\aAuxAll}{14.90}

\providecommand{\aResMcar}{4.45}\providecommand{\aResMar}{10.34}
\providecommand{\aResNmar}{27.38}\providecommand{\aResAll}{14.06}

\providecommand{\aAtnMcar}{4.15}\providecommand{\aAtnMar}{10.22}
\providecommand{\aAtnNmar}{27.14}\providecommand{\aAtnAll}{13.84}

\providecommand{\aLinMcar}{2.91}\providecommand{\aLinMar}{12.34}
\providecommand{\aLinNmar}{28.94}\providecommand{\aLinAll}{14.73}

\providecommand{\aDAux}{$+2.74$}
\providecommand{\aDRes}{$+1.90$}
\providecommand{\aDAtn}{$+1.68$}
\providecommand{\aDLin}{$+2.57$}

%% file: 4b_ablation.tex
\subsection{Ablation Studies}
\label{sec:exp-ablation}

Two choices separate CAIR\ from a generic masked autoencoder: the base curve is
learned \emph{and supervised on the interpolation task}, and it reaches the
refiner as an additive residual rather than as conditioning alone.
We remove each in turn, retrain from scratch, and score on the protocol of
Table~\ref{tab:mech}, so the full model and linear interpolation carry over
unchanged (Table~\ref{tab:ablation}).

All three ablations reduce accuracy, and their ordering identifies the component
that contributes most.
Removing the auxiliary interpolation loss is the most expensive
(\aDAux\,mg/dL on the three-mechanism mean, $+22.5\%$), the residual path next
(\aDRes), and replacing the bidirectional GRU with a self-attention block, the
estimate-complete-re-estimate structure of SAITS \citep{saits}, the
least (\aDAtn).
What separates CAIR\ from that family is thus not that its first stage is
neural, nor which sequence model implements it, but that the stage is supervised
on the interpolation task itself, the property Sec.~\ref{sec:related} identifies
as absent from generic imputers: without that term the model is worse on the
mean than linear (\aAuxAll\ vs.\ \aLinAll).

The effect is not uniform. Under MAR the four model rows lie within
$0.53$\,mg/dL and the attention variant is nominally ahead, so there the
components are near-interchangeable; the mechanisms that separate them are MCAR,
where the base curve is nearly sufficient and every ablated variant is less
accurate than linear interpolation, and NMAR, where they give up
$3.6$--$4.1$\,mg/dL but all remain more accurate than it. Cells are
single runs, so the full row is the reference and we do not read the MAR spread
as an effect.
The third component reflected in the name of the method, the training
curriculum, is ablated separately in Sec.~\ref{sec:exp-mimic}: replacing the
broad curriculum with masks tailored to the physiology of a source domain,
architecture unchanged, makes the model less accurate than linear interpolation.

\begin{table}[htpb]
  \centering
  {\small
  \setlength{\tabcolsep}{3pt}
  \begin{tabular}{lccccr}
    \toprule
    Variant & MCAR & MAR & NMAR & \textbf{ALL} $\downarrow$ & $\Delta$ \\
    \midrule
    \textbf{CAIR\ (full)}    & \textbf{\aFullMcar} & \underline{\aFullMar} & \textbf{\aFullNmar} & \textbf{\aFullAll} & n/a \\
    \midrule
    \; w/o aux.\ loss         & \aAuxMcar & \aAuxMar & \aAuxNmar & \aAuxAll & \aDAux \\
    \; w/o residual           & \aResMcar & \aResMar & \aResNmar & \aResAll & \aDRes \\
    \; GRU $\to$ self-attn.   & \aAtnMcar & \textbf{\aAtnMar} & \underline{\aAtnNmar} & \underline{\aAtnAll} & \aDAtn \\
    \midrule
    linear interp             & \underline{\aLinMcar} & \aLinMar & \aLinNmar & \aLinAll & \aDLin \\
    \bottomrule
  \end{tabular}}
  \caption{\textbf{Component ablation} on AI-READI CGM, on the protocol of
  Table~\ref{tab:mech} (all $\aNpart$ test participants; RMSE mg/dL over six
  missingness rates). \textbf{ALL} is the mean of the three mechanisms,
  $\Delta$ the change against the full model. Rows are single training runs
  under the published recipe, differing only in the ablated component; linear
  is repeated from Table~\ref{tab:mech} for scale.
  \textbf{Bold} = best, \underline{underline} = second best, per metric column.}
  \label{tab:ablation}
\end{table}

%% file: 5_conclusion.tex
\section{Conclusion}
\label{sec:conclusion}
We address the gap highlighted by \citet{toye}, where linear interpolation
outperforms every learned imputer on clinical time series with realistic gaps.
CAIR\ is a two-stage \emph{interpolate-then-refine} model: a learned
interpolator predicts a base curve inside each gap, and an iteratively unrolled
Transformer refines it, with both stages trained jointly under a broad gap
curriculum.
The design targets the two properties of physiological missingness that generic
imputers ignore: mechanism-driven gaps (C1) and gap lengths spanning orders of
magnitude (C2).

CAIR\ is the most accurate method under every missingness mechanism
on both glucose and arterial pressure, its margin growing with difficulty to
$19\%$ under value-dependent dropout.
The generic learned imputers that fail in \citet{toye} fail here too, placing
the cause in the training curriculum rather than neural capacity.
We verify this by changing only the gap distribution, which on its own removes
the advantage over linear interpolation.
Reconstruction accuracy and clinical-metric fidelity are distinct axes, and
CAIR\ alone ranks among the best on both.

%% file: 99_acknowledgement.tex
\section*{Acknowledgment}
This research was partially funded by the National Institutes of Health (NIH) under award
1OT2OD038051. The views and conclusions contained in this document are those of the authors
and should not be interpreted as representing the official policies, either expressed or implied, of the
NIH.

%% file: 6_appendix.tex
\section{Extended Related Work}
\label{sup:related}

\subsection{Interpolation and smoothing}
Deployed clinical pipelines overwhelmingly use deterministic interpolants.
Linear interpolation draws a chord between the observations bracketing a gap; it
is exactly reconstructive when the underlying signal is locally affine and
degrades gracefully otherwise, which explains its persistent strength on smooth
vitals.
Shape-preserving cubics improve on it by constraining the interpolant's
derivatives: PCHIP \citep{pchip} enforces monotonicity on monotone data, so it
does not introduce spurious overshoot at the edges of a gap, and the Akima
spline \citep{akima} computes slopes from a local five-point stencil, which
makes it robust to outliers at the cost of second-derivative continuity.
Unconstrained cubic splines \citep{cubicspline} are smoother still but overshoot
badly across long gaps, which is visible in our physiological protocol where
cubic spline is the second-worst method overall.
Savitzky--Golay filtering \citep{savgol} fits a low-order polynomial in a
sliding window by least squares and is a smoother rather than an interpolator;
it is standard in CGM preprocessing.
Exponentially weighted moving averages \citep{ewma} and Kalman smoothing
\citep{kalman} bring an explicit state model, but both assume a stationarity
that physiological signals violate across meals and sleep.

\subsection{Statistical and tabular imputation}
A second family treats the series as a table of correlated features.
Multiple imputation by chained equations \citep{mice} iteratively regresses each
variable on the others and produces proper multiple imputations, so it carries
uncertainty correctly under MAR.
missForest \citep{missforest} replaces those conditional models with random
forests, which captures interactions without a parametric specification.
$k$-nearest-neighbour imputation \citep{knn_impute} fills a value from the most
similar complete records, and hot-deck imputation \citep{hotdeck} donates
observed values from a matched donor rather than synthesizing them.
All four recover distributional structure that interpolation discards.
Their weakness on time series is the mirror image: because they do not model
temporal order, their reconstructions are not smooth, which is why
they score well on threshold-based burden metrics and poorly on RMSE in
Table~4 of the main paper.
Last-observation-carried-forward is the degenerate member of this family and its
biases are well documented \citep{locf}.

\subsection{Learned sequence imputation}
Recurrent imputers exploit informative missingness directly.
GRU-D \citep{gru_d} adds a learned decay that pulls the hidden state toward the
empirical mean as the time since the last observation grows.
BRITS \citep{brits} imputes in both directions and penalizes disagreement
between them, treating missing values as trainable variables.
M-RNN \citep{mrnn} combines within-stream interpolation and across-stream
imputation in a multi-directional architecture.
Attention-based and generative models followed.
SAITS \citep{saits} is the closest published architecture to CAIR.
It runs a diagonally-masked self-attention (DMSA) block to obtain a first
estimate, replaces the missing entries of the input with that estimate to form a
completed series, passes the completed series to a second DMSA block, and blends
the two blocks' outputs through a gate computed from the attention map and the
missingness mask; its reconstruction loss is accumulated over the first block's
output, the second block's output, and the blend, so the intermediate estimate
is supervised.
CAIR\ shares this estimate-complete-re-estimate structure and differs in every
stage of it.
Its first stage is a bidirectional GRU restricted to the observed values and the
mask rather than a second attention block, and it is supervised at the
\emph{held-out} positions, so it is trained on the interpolation task itself
rather than on reconstructing values it can already see.
Its second stage receives the base curve as an additive residual as well as
through the conditioning, which makes the refiner's regression target the
residual by construction.
And a single refiner is applied with tied weights rather than two distinct
blocks, so the number of passes is a deployment choice rather than an
architectural constant (we use three at training and at inference), and each
pass is told, through a three-valued provenance
code, which entries are observations and which are its own earlier fills.
GP-VAE \citep{gpvae} places a Gaussian-process prior over the latent trajectory
of a VAE, giving calibrated uncertainty.
CSDI \citep{csdi} formulates imputation as conditional score-based diffusion,
and general-purpose backbones such as TimesNet \citep{timesnet} treat imputation
as one task among several.

These models are designed to be domain-agnostic. Both M-RNN and GP-VAE fall
below every non-constant baseline on \emph{both} of our domains, reproducing the
result of \citet{toye}. They differ from CAIR\ in the gap distribution they are
trained under rather than in capacity.

\subsection{Evaluating under realistic missingness}
Rubin's MCAR/MAR/NMAR taxonomy \citep{rubin} is standard in statistics but is
rarely used to structure machine-learning imputation benchmarks, which typically
delete values completely at random and report one averaged error.
\citet{beyond_random_missingness} argue that clinically plausible missingness
patterns change method rankings in healthcare time series, and
\citet{icu_imputation_bench} reach a similar conclusion for critical care.
\citet{cgm_missing_metrics} show specifically that the choice of imputation
strategy changes the CGM metrics clinicians read.
Closest to our work, \citet{toye} evaluate eleven imputers on real-world clinical
signals under mechanism-driven deletion and find that linear interpolation
outperforms all of them. We adopt their protocol verbatim as our hardest
evaluation.

\section{Detailed Experimental Setup}
\label{sup:setup}

\subsection{Data preparation}
\paragraph{AI-READI.}
We use the flagship type-2 diabetes release \citep{aireadi,aireadi_dataset}.
CGM traces are resampled onto a uniform 5-minute grid; heart rate and
respiration are resampled onto the same grid from the wearable's native
sampling. Participants are split $1{,}576$ / $352$ / $352$ into train,
validation and test by participant identifier, so no participant appears in two
splits. The evaluation target is day two of each trace, indices $[288,576)$, a
$24$\,h window. Values are $z$-normalized with the training mean and standard
deviation ($\mu = 132.05$, $\sigma = 42.33$\,mg/dL for CGM) and all errors are
rescaled to native units for reporting.

\paragraph{MIMIC-III.}
We extract $22{,}156$ twenty-four-hour windows on a 5-minute grid from the
numerics of MIMIC-III \citep{mimic3}, retaining the observed sensor masks rather
than imposing complete data. Splits are subject-disjoint. Targets
are mean arterial blood pressure and heart rate; the covariate channel used for
the MAR trigger is a co-recorded vital distinct from the target.

\subsection{Missingness mechanisms}
Each mechanism is applied to the target window at six missingness rates
($5, 10, 15, 20, 25, 30\%$), with five seeded masks per rate.
\emph{MCAR} deletes positions independently at the target rate.
\emph{MAR} deletes contiguous windows whose placement is drawn from a
distribution over a co-recorded covariate: on AI-READI the time-aligned wearable
activity signal, available for $303$ of the $352$ test participants; on
MIMIC-III a co-recorded covariate vital. The target value itself never enters
the trigger.
\emph{NMAR} deletes contiguous windows triggered by the target's own value
crossing a clinically extreme threshold ($<70$ or $>150$\,mg/dL for glucose;
the analogous clinical bands for ABP and HR).

\subsection{CAIR\ configuration}
Stage~1 is a 4-layer bidirectional GRU with hidden size $128$ and a
zero-initialized linear output head. Stage~2 is an 8-layer pre-norm Transformer
encoder with $d_{\text{model}} = 128$, $8$ attention heads and FFN width $512$,
operating bidirectionally with no causal mask. Each position is encoded as the
sum of a linear embedding of the observed value (a learned mask token at missing
positions), learned day and time-of-day embeddings over the $288$ five-minute
bins of a day, an embedding of the observation state, and a linear projection of
the $42$-dimensional feature vector $\phi$. The layout of $\phi$ is: diurnal
history ($14$), window-level summary statistics ($5$), gap geometry ($2$), the
two interpolant value slots and their validity flags ($4$), boundary first
differences ($4$), boundary second differences ($4$), least-squares boundary
slopes ($4$), and gap context ($5$). Four of those dimensions form an interpolant
interface: a conventional pipeline fills them with a linear and a PCHIP estimate
of the missing value, each with a flag marking it present. CAIR\ computes
neither. It writes the Stage-1 prediction $y^{0}$ into both value slots and sets
both flags, so the refiner reads one learned base curve where it would otherwise
read two deterministic ones.

Training uses AdamW \citep{adamw} at learning rate $3\times10^{-4}$ with cosine
decay, batch size $64$, and an exponential moving average of the weights with
decay $0.999$ used for inference \citep{ema_swa}. The refiner is unrolled for
three passes with loss weights $(0.15, 0.35, 0.50)$ and the Stage-1 auxiliary
loss is weighted $\lambda_{\text{aux}} = 0.7$. The random-gap curriculum holds
out approximately $20\%$ of observed samples per step as a mixture of scattered
points and contiguous blocks. The block-length distribution is signal-agnostic:
it is specialized to no domain in particular, and it never coincides with an
evaluation mask.

At inference we slide a $576$-step window with stride $144$, average overlapping
predictions under a cosine window, and run the same three refiner passes used in
training, a base pass plus two conditioned on the previous estimate. Predictions
at observed positions are left unchanged. Five seeds are averaged into a deep
ensemble \citep{deep_ensemble}. Each member trains on a single NVIDIA
RTX~6000~Ada GPU.

\subsection{Baseline configuration}
All baselines are evaluated on the identical masks and seeds as CAIR.
Interpolants and smoothers use the SciPy implementations
(\texttt{interp1d}, \texttt{PchipInterpolator}, \texttt{Akima1DInterpolator},
\texttt{CubicSpline}, \texttt{savgol\_filter}) at their default settings;
Savitzky--Golay uses a window of $31$ samples and polynomial order $3$; EWMA
uses a smoothing factor $\alpha = 0.3$, applied in both directions and averaged
where the forward and backward estimates overlap; the Kalman smoother uses a
local-linear-trend
state model fitted per trace. The \emph{mode} baseline fills each gap with the
most frequent observed value in the window. On the physiological protocol the
\emph{mean} fill is applied locally, over a centered $48$-sample ($4$\,h)
neighborhood rather than the whole trace, falling back to the window-level mean
where that neighborhood contains no observation; we write it \emph{local mean}
in Table~\ref{tab:phys-rest} to distinguish it from the trace-level mean fill of
Table~1 of the main paper.
The bidirectional AR baseline fits an order-$12$
autoregressive model forward and backward and blends the two predictions
linearly across the gap. Tabular imputers (MICE, missForest, $k$NN, hot-deck) treat
each window as a feature vector; $k$NN uses $k=5$ with a masked Euclidean
metric and hot-deck is its $k=1$ donor limit. SAITS, BRITS, M-RNN and GP-VAE are
trained on the same windows and the same curriculum budget as CAIR, using the
authors' published hyperparameters where available.

\subsection{Which baselines run on which protocol}
\label{sup:subsets}
The four protocols score overlapping but not identical baseline subsets, and we
state the mapping here.
The three-mechanism protocol is the broadest and is what establishes the
ordering between families: it scores the constant and interpolation fills
(linear, LOCF, mean), the spectral reconstruction (Fourier), all four tabular
imputers (MICE, missForest, $k$NN, hot-deck) and both generic learned imputers
(M-RNN, GP-VAE). The MIMIC-III transfer repeats that same set on arterial
pressure and heart rate (Sec.~\ref{sup:mimic-rest}).
The gap-length and physiological protocols instead target the short- and
structured-gap regimes, where the ordering established above makes the constant
fills and tabular imputers uninformative: they score the interpolants and
smoothers that are competitive there (PCHIP, akima, cubic spline,
Savitzky--Golay, EWMA, Kalman, and on heart rate and respiration a
bidirectional AR), together with SAITS and BRITS, the two learned imputers
architecturally closest to CAIR.
The rows the main paper abbreviates are reported in Sec.~\ref{sup:phys-rest} and
Sec.~\ref{sup:mimic-rest}.

\subsection{Metrics}
RMSE is computed over held-out positions only, in native units. The
metric-recovery ratio for a clinical metric $g$ is
\begin{equation}
\mathrm{MRR}(g) = 1 - \frac{\lvert g(\hat{x}) - g(x)\rvert}
                          {\lvert g(x_{\text{mean}}) - g(x)\rvert},
\end{equation}
where $x_{\text{mean}}$ is the mean-filled trace, so $\mathrm{MRR}=1$ is exact
recovery, $0$ is no better than mean-fill, and negative values are worse than
mean-fill. Clinical metrics are time-in-range \citep{battelino_tir}, MAGE
\citep{mage}, coefficient of variation, and for MIMIC-III the fractions of time
in, above and below the normal band.

\section{External-Cohort Pretraining Does Not Improve Accuracy}
\label{sup:pretrain}

Pretraining on external CGM cohorts does not improve accuracy on the target
domain. Three pools spanning both type-1 and type-2 physiology leave the
physiological average within $0.2$\,mg/dL of a model trained on AI-READI alone,
and none improves on it.

\paragraph{Procedure.}
External datasets are standardized to the AI-READI schema through a
dataset-adapter registry and pooled with a domain-balanced sampler. A
zero-initialized per-dataset \emph{domain embedding} $e_d$ is added to the
Transformer's per-position conditioning. We pretrain on
AI-READI $\cup$ external, then fine-tune on AI-READI alone with $e_0$ frozen at
zero, so that inference is bit-identical to a model that never saw external
data. Any gain must therefore survive in-domain fine-tuning.

\paragraph{Result.}
Table~\ref{tab:pretrain} reports the physiological protocol for three
pretraining pools: HUPA-UCM \citep{hupaucm} ($18$ type-1 adults), OhioT1DM
\citep{ohiot1dm} ($6$ type-1 adults) and Shanghai T2DM \citep{shanghai}
($109$ type-2 adults). Every external cell lands within $0.2$\,mg/dL of the
in-domain control and none below it. The distribution the pool is drawn from
does not change this: off-distribution (type-1) and on-distribution (type-2)
pretraining behave alike, and fine-tuning on a $1{,}576$-participant target
recovers the same minimizer to within $0.2$\,mg/dL regardless of which pool
preceded it.

\begin{table}[t]
  \centering
  {\small
  \setlength{\tabcolsep}{2.4pt}
  \begin{tabular}{lcccccc}
    \toprule
    Pretrain, then fine-tune & meal & sleep & asc & dip & comb & AVG \\
    \midrule
    control (AI-READI only)  & \textbf{15.60} & 27.52 & \textbf{5.96} & \textbf{5.14} & \textbf{14.74} & \textbf{13.79} \\
    $+$ HUPA-UCM (T1D)       & 15.99 & \textbf{26.32} & 6.21 & 5.36 & 15.66 & 13.91 \\
    $+$ OhioT1DM (T1D)       & 15.77 & \underline{27.14} & \underline{6.08} & 5.21 & 15.25 & \underline{13.89} \\
    $+$ Shanghai (T2D)       & \underline{15.71} & 27.58 & 6.13 & \underline{5.20} & \underline{15.18} & 13.96 \\
    \bottomrule
  \end{tabular}}
  \caption{\textbf{External-cohort pretraining does not improve accuracy}
  (physiological protocol, $352$ participants, single seed; RMSE mg/dL). Every
  external cell is within $0.2$\,mg/dL of the control and none improves on it.
  \textbf{Bold} = best, \underline{underline} = second best, per column.}
  \label{tab:pretrain}
\end{table}

The same conclusion holds on the gap-length protocol: the Shanghai-pretrained
checkpoint scores $7.73$\,mg/dL on the short-gap mean against $7.99$ for its own
control, both far worse than the $5.69$ of the main model, so the pretraining
delta is negligible next to the effect of the training curriculum.

\paragraph{Why this control's average is lower than the published model's.}
The control row of Table~\ref{tab:pretrain} attains $13.79$\,mg/dL on the
physiological average, lower than the $14.61$ of the published model in Table~3
of the main paper. The two use different recipes: this experiment uses a
two-stage recipe selected against the physiological average, and that
average is governed by the multi-hour \emph{sleep} blocks
($26$--$28$\,mg/dL), which dwarf the short \emph{ascending} and \emph{dipping}
segments ($5$--$6$\,mg/dL). Selecting a checkpoint against it therefore rewards
handling large masked fractions and over-smoothing the short gaps: the same
recipe is more than $2$\,mg/dL worse at \emph{every} gap length of Table~2 of
the main paper ($7.99$ against $5.69$ on the short-gap mean, above). All rows of
Table~\ref{tab:pretrain} use this recipe, so the comparison across pretraining
pools is unaffected. This is the same effect (C2) describes in the main paper: a
single averaged score, dominated by an unrealistically long-gap regime, can rank
a model first while it is worse in the regime clinical practice requires.

\section{Additional Results}
\label{sup:extra}

\subsection{Remaining baselines on the physiological protocol}
\label{sup:phys-rest}
Table~3 of the main paper reports the physiological five-strategy protocol over
the baselines that are competitive on it. Table~\ref{tab:phys-rest}
completes that table with the four weakest, which were omitted there for space:
EWMA, a local mean, an unconstrained cubic spline and LOCF.
The weakest method carried in the
main table, Savitzky--Golay, averages $18.15$\,mg/dL; the best of the four here
averages $23.03$ and the worst $31.72$, so the gap from the main table's floor to
this group ($4.9$\,mg/dL) is larger than the gap from CAIR to that floor
($3.5$\,mg/dL). Two failure modes separate them. The cubic spline is accurate on
the short \emph{ascending} and \emph{dipping} segments but overshoots badly
across the long \emph{sleep} blocks ($45.22$\,mg/dL, the worst cell in the
table), which is the known cost of an unconstrained third-order fit over a wide
gap and the reason shape-preserving cubics such as PCHIP and akima are preferred
in deployment. LOCF is the reverse: holding the last observation is uniformly
poor and degrades most on the short segments ($34.19$ and $29.36$\,mg/dL), where
a fill is expected to track a rising or falling trend rather than freeze it.

\begin{table}[t]
  \centering
  {\small
  \setlength{\tabcolsep}{3.2pt}
  \begin{tabular}{lcccccc}
    \toprule
    Method & meal & sleep & asc & dip & comb & \textbf{AVG} $\downarrow$ \\
    \midrule
    \textbf{CAIR\ (Ours)} & \textbf{15.25} & \textbf{24.94} & \textbf{9.01} & \textbf{7.49} & \textbf{16.37} & \textbf{14.61} \\
    Savitzky--Golay & \underline{17.27} & \underline{29.45} & \underline{13.41} & \underline{11.22} & \underline{19.43} & \underline{18.15} \\
    \midrule
    EWMA          & 20.59 & 31.69 & 21.45 & 18.32 & 23.11 & 23.03 \\
    local mean    & 22.66 & 32.23 & 24.00 & 20.65 & 24.40 & 24.79 \\
    cubic spline  & 22.64 & 45.22 & 21.90 & 18.62 & 23.85 & 26.44 \\
    LOCF          & 25.91 & 37.33 & 34.19 & 29.36 & 31.81 & 31.72 \\
    \bottomrule
  \end{tabular}}
  \caption{\textbf{The four baselines omitted from Table~3 of the main
  paper} (AI-READI, physiological five-strategy protocol, all $352$ test
  participants; RMSE mg/dL, lower is better). The lower block is the omitted
  group; CAIR\ and Savitzky--Golay are repeated from Table~3 as the best method
  and the weakest one carried there. Every omitted method is at least
  $4.9$\,mg/dL behind that floor on the average. \textbf{Bold} = best,
  \underline{underline} = second best, per column.}
  \label{tab:phys-rest}
\end{table}

\subsection{Conditioning-modality ladder on CGM}
Table~\ref{tab:ladder} sweeps the conditioning set for CAIR\ on AI-READI CGM,
from the target channel alone up to ten co-recorded modalities, five seeds per
rung. The physiological average moves from $13.11$ to $12.98$\,mg/dL across the
entire ladder and is non-monotone, so no rung is distinguishable from the
control at this seed count. This is what makes the heart-rate and respiration
result in the main paper informative: the identical conditioning mechanism adds
nothing on glucose and $20$--$34\%$ on the exogenously driven signals, so what
transfers is the mechanism's dependence on the target, not the mechanism itself.

\begin{table}[t]
  \centering
  {\small
  \setlength{\tabcolsep}{4pt}
  \begin{tabular}{lccccc}
    \toprule
    Conditioning set & meal & sleep & asc & dip & \textbf{AVG} \\
    \midrule
    target only (control) & 14.94 & 25.17 & 6.07 & 4.93 & 13.11 \\
    $+$ heart rate        & 14.93 & 25.16 & 6.03 & 4.92 & 13.10 \\
    $+$ steps, calories   & 14.89 & 25.25 & 6.10 & 4.97 & 13.14 \\
    $+$ sleep state       & 14.89 & 25.27 & 5.98 & 4.92 & 13.09 \\
    $+$ respiration, stress & 14.79 & 24.96 & 6.00 & 4.89 & \textbf{12.98} \\
    $+$ environment       & 14.82 & 24.83 & 6.24 & 5.06 & 13.10 \\
    $+$ clinical          & 14.79 & 24.85 & 6.18 & 5.01 & \underline{13.06} \\
    $+$ ECG               & 14.86 & 24.87 & 6.26 & 5.09 & 13.14 \\
    $+$ retinal           & 14.86 & 24.90 & 6.12 & 4.98 & 13.07 \\
    \bottomrule
  \end{tabular}}
  \caption{Conditioning-modality ladder on AI-READI CGM (five-seed ensemble per
  rung; RMSE mg/dL). Adding modalities does not improve glucose imputation; the AVG spread ($12.98$--$13.14$) is within seed noise. AVG is over all five physiological strategies; the combined column is omitted for space. Absolute values are not comparable to Table~3 of the main paper. \textbf{Bold}/\underline{underline} mark the best/second AVG.}
  \label{tab:ladder}
\end{table}

\subsection{Complete MIMIC-III baselines, both signals}
\label{sup:mimic-rest}
Table~6 of the main paper reports CAIR, linear interpolation and the two generic
learned imputers on arterial pressure and heart rate.
Tables~\ref{tab:mimic-full} and \ref{tab:mimic-hr-full} complete it with both
CAIR\ variants and the constant-fill and tabular baselines omitted there for
space. On arterial pressure both CAIR\ variants are more accurate than every
baseline under every mechanism.
On heart rate CAIR\ and linear interpolation are separated by at most
$0.22$\,bpm under any mechanism, so the main paper reports the signal as a
boundary case. That boundary is between those two methods alone: under MAR and
NMAR the closest remaining baseline, LOCF, still trails linear interpolation by
$1.6$ and $1.4$\,bpm. Under MCAR the whole field is tight (missForest is within
$0.15$\,bpm of linear), consistent with scattered single-sample deletion on a
smooth signal.

\begin{table}[t]
  \centering
  {\small
  \setlength{\tabcolsep}{6pt}
  \begin{tabular}{lccc}
    \toprule
    Method & MCAR & MAR & NMAR \\
    \midrule
    CAIR\ (univariate)   & \textbf{4.26} & \textbf{8.58} & \underline{11.44} \\
    CAIR\ (multivariate) & \underline{4.30} & \underline{8.60} & \textbf{11.42} \\
    linear interp   & 4.40 & 9.23 & 11.85 \\
    missForest      & 4.62 & 9.60 & 12.91 \\
    MICE            & 4.88 & 9.02 & 12.65 \\
    $k$NN           & 5.27 & 8.91 & 12.69 \\
    hot-deck        & 5.82 & 9.11 & 12.81 \\
    LOCF            & 5.89 & 10.43 & 12.79 \\
    Fourier         & 5.94 & 11.28 & 13.56 \\
    mean            & 9.33 & 9.93 & 14.01 \\
    mode            & 11.42 & 11.82 & 15.11 \\
    GP-VAE          & 11.03 & 12.82 & 17.94 \\
    M-RNN           & 13.45 & 13.19 & 18.35 \\
    \bottomrule
  \end{tabular}}
  \caption{Complete MIMIC-III arterial-pressure reconstruction RMSE (mmHg, mean
  over six missingness rates). \textbf{Bold} = best, \underline{underline} = second best, per column.}
  \label{tab:mimic-full}
\end{table}

\begin{table}[t]
  \centering
  {\small
  \setlength{\tabcolsep}{6pt}
  \begin{tabular}{lccc}
    \toprule
    Method & MCAR & MAR & NMAR \\
    \midrule
    CAIR\ (univariate)   & \underline{2.26} & \underline{5.36} & 6.72 \\
    CAIR\ (multivariate) & \underline{2.26} & \textbf{5.28} & \underline{6.48} \\
    linear interp   & \textbf{2.22} & 5.50 & \textbf{6.42} \\
    missForest      & 2.37 & 7.40 & 9.25 \\
    MICE            & 2.49 & 7.55 & 9.63 \\
    $k$NN           & 2.69 & 7.09 & 9.21 \\
    hot-deck        & 2.90 & 7.19 & 9.28 \\
    LOCF            & 3.07 & 7.06 & 7.79 \\
    Fourier         & 3.31 & 7.17 & 7.86 \\
    mean            & 8.41 & 9.22 & 11.69 \\
    mode            & 10.88 & 10.39 & 14.10 \\
    GP-VAE          & 9.17 & 14.01 & 18.00 \\
    M-RNN           & 15.28 & 14.53 & 18.60 \\
    \bottomrule
  \end{tabular}}
  \caption{Complete MIMIC-III heart-rate reconstruction RMSE (bpm, mean over six
  missingness rates). The CAIR\ (multivariate) and linear rows are those of
  Table~6 of the main paper. CAIR\ is the most accurate method under MAR and
  linear interpolation under MCAR and NMAR, the two separated by at most
  $0.22$\,bpm; the per-mechanism significance tests for those differences are in
  Table~\ref{tab:mimic-wilcoxon}.
  \textbf{Bold} = best, \underline{underline} = second best, per column.}
  \label{tab:mimic-hr-full}
\end{table}

\subsection{Significance tests on MIMIC-III}
Table~\ref{tab:mimic-wilcoxon} reports the paired Wilcoxon signed-rank test of
CAIR\ (multivariate) against linear interpolation, computed per
(window, rate) pair and reported separately for each signal and mechanism.
On arterial pressure all three tests favor CAIR, the weakest at
$p=2.9\times10^{-3}$, which is the bound quoted in Sec.~4.6 of the main paper.
On heart rate the picture is mixed and we report it in full: linear
interpolation is significantly more accurate under MCAR, while the MAR and NMAR
differences are not significant at any conventional level, so on those two
mechanisms the two methods are statistically tied.

\begin{table}[t]
  \centering
  {\small
  \setlength{\tabcolsep}{6pt}
  \begin{tabular}{llcc}
    \toprule
    Signal & Mechanism & $\Delta$ & $p$ \\
    \midrule
    \multirow{3}{*}{ABP-mean (mmHg)}
      & MCAR & $-0.105$ & $7.9\times10^{-9}$ \\
      & MAR  & $-0.625$ & $4.1\times10^{-21}$ \\
      & NMAR & $-0.375$ & $2.9\times10^{-3}$ \\
    \midrule
    \multirow{3}{*}{Heart rate (bpm)}
      & MCAR & $+0.040$ & $7.0\times10^{-40}$ \\
      & MAR  & $-0.220$ & $0.83$ \\
      & NMAR & $+0.060$ & $0.37$ \\
    \bottomrule
  \end{tabular}}
  \caption{\textbf{Paired Wilcoxon signed-rank tests on MIMIC-III}, CAIR\
  (multivariate) versus linear interpolation, per (window, rate) pair.
  $\Delta$ is the mean of the per-pair RMSE differences in native units, which
  need not equal the difference of the aggregate RMSEs in Table~6 of the main
  paper because RMSE does not aggregate linearly; negative favors CAIR. All three arterial-pressure tests favor CAIR; on heart rate,
  linear interpolation is significantly better under MCAR and the remaining two
  differences are not significant ($\alpha=0.05$).}
  \label{tab:mimic-wilcoxon}
\end{table}

\subsection{Reconstruction vs.\ burden on MIMIC-III ABP}
Table~\ref{tab:mimic-down} gives the full method-by-method breakdown of the two
axes summarized in the main paper: interpolants attain low RMSE but fail to
preserve the clinical burden, tabular and neural imputers recover the burden at
much higher RMSE, and only CAIR\ occupies both corners.

\begin{table}[t]
  \centering
  {\small
  \setlength{\tabcolsep}{4pt}
  \begin{tabular}{lcc}
  \toprule
  Method & RMSE (mmHg) $\downarrow$ & Burden MRR $\uparrow$ \\
  \midrule
  \multicolumn{3}{l}{\emph{Low RMSE, burden not preserved}} \\
  \quad linear interp & 11.85 & $+0.14$ \\
  \quad LOCF & 12.79 & $+0.07$ \\
  \quad Fourier & 13.56 & $+0.05$ \\
  \midrule
  \multicolumn{3}{l}{\emph{Recovers burden, high RMSE}} \\
  \quad MICE & 12.65 & $+0.63$ \\
  \quad $k$NN & 12.69 & $\mathbf{+0.65}$ \\
  \quad hot-deck & 12.81 & $\mathbf{+0.65}$ \\
  \quad GP-VAE & 17.94 & $+0.64$ \\
  \quad M-RNN & 18.35 & $+0.64$ \\
  \midrule
  \multicolumn{3}{l}{\emph{Both}} \\
  \quad CAIR\ (univariate) & \underline{11.44} & $\mathbf{+0.65}$ \\
  \quad \textbf{CAIR\ (multivar.)} & \textbf{11.42} & $+0.64$ \\
  \bottomrule
  \end{tabular}}
  \caption{\textbf{Only CAIR\ ranks among the best on both axes} (MIMIC-III ABP,
  NMAR, mean over six rates). Reconstruction RMSE (mmHg, lower better) and
  clinical-burden recovery (MRR on time in/above/below range, higher better).
  Interpolants match the RMSE of CAIR\ but fail to preserve the burden; tabular and neural
  imputers recover the burden at much higher RMSE, with $k$NN and hot-deck
  matching its recovery at ${\sim}11\%$ higher RMSE. \textbf{Bold} = best, \underline{underline} = second best, per column.}
  \label{tab:mimic-down}
\end{table}

\subsection{Single-vital outcome prediction}
\label{sup:singlevital}
Sec.~4.7 of the main paper reports that hard stay-level outcomes are insensitive
to reconstruction fidelity. Table~\ref{tab:singlevital} reports that experiment.
We isolate the effect of the fill by predicting in-hospital mortality from
arterial pressure \emph{alone}, so that the imputed channel is the classifier's
only input and cannot be compensated by co-recorded vitals. The readout is a
1D-CNN over the completed trace, which is sensitive to the shape of the
reconstruction rather than to summary statistics of it. The oracle row is the
recorded trace with no deletion applied, and is therefore the ceiling this task
can reach.

No fill is distinguishable from the oracle, and four of the five exceed it.
Under MCAR every imputer improves on mean-fill, as expected, but linear
interpolation and PCHIP then score $0.692$ against the oracle's $0.679$: a
reconstruction further from the recorded signal yields a
\emph{better} classifier than the recorded signal itself. The mechanism is that
interpolation denoises the vital, and a weak-signal classifier rewards
smoothing; under NMAR, where the deletions sit on the clinically extreme
excursions, the inversion is no smaller and becomes uniform, with every fill
including mean-fill above the oracle.
The fidelity CAIR\ optimizes is therefore orthogonal to
what this task rewards. We report reconstruction error and burden recovery
(Table~\ref{tab:mimic-down}) rather than hard-outcome AUROC. MAR is omitted
because the outcome experiment was run under MCAR and NMAR only.

\begin{table}[t]
  \centering
  {\small
  \setlength{\tabcolsep}{6pt}
  \begin{tabular}{lcc}
    \toprule
    Fill & MCAR AUROC & NMAR AUROC \\
    \midrule
    oracle (recorded trace) & 0.679 & 0.678 \\
    \midrule
    mean-fill      & 0.677 & 0.683 \\
    $k$NN          & 0.680 & 0.685 \\
    CAIR\ (multivar.) & 0.688 & 0.688 \\
    linear interp  & 0.692 & 0.688 \\
    PCHIP          & 0.692 & 0.689 \\
    \bottomrule
  \end{tabular}}
  \caption{\textbf{Hard outcomes do not reward reconstruction fidelity}
  (MIMIC-III, in-hospital mortality predicted from arterial pressure alone,
  $30\%$ missingness, 1D-CNN readout; $n=1200$ under MCAR and $1181$ under
  NMAR). The oracle is the recorded trace and is the ceiling for this task, yet
  four fills exceed it under MCAR and all five do under NMAR, because
  interpolation denoises the vital. No column is ordered as reconstruction
  accuracy would predict, and no best value is marked.}
  \label{tab:singlevital}
\end{table}

\subsection{Per-cohort stratification on CGM}
Under NMAR, CAIR\ attains both lower RMSE and higher time-in-range recovery than
linear interpolation in all four AI-READI diabetes study groups
(Table~\ref{tab:cohort}). The advantage is present in the healthy
group, where absolute error is lowest, and largest in the insulin-dependent
group, where the excursions that NMAR deletes are most frequent.
AI-READI de-identifies gender, so study group is the available and more
physiologically relevant stratification axis.

\begin{table}[t]
  \centering
  {\small
  \setlength{\tabcolsep}{4pt}
  \begin{tabular}{lcccc}
    \toprule
    & \multicolumn{2}{c}{RMSE (mg/dL) $\downarrow$} & \multicolumn{2}{c}{TIR recovery $\uparrow$} \\
    \cmidrule(lr){2-3}\cmidrule(lr){4-5}
    Study group & CAIR & linear & CAIR & linear \\
    \midrule
    healthy            & \textbf{18.29} & 21.49 & \textbf{0.165} & 0.097 \\
    pre-diabetes       & \textbf{15.90} & 21.15 & \textbf{0.467} & 0.382 \\
    oral medication    & \textbf{33.96} & 40.19 & \textbf{0.601} & 0.556 \\
    insulin-dependent  & \textbf{27.23} & 35.17 & \textbf{0.608} & 0.489 \\
    \bottomrule
  \end{tabular}}
  \caption{\textbf{The CGM result holds in every diabetes study group}
  (AI-READI, NMAR, all $352$ test participants). CAIR\ attains lower
  reconstruction error and higher time-in-range recovery than linear
  interpolation in all four groups. \textbf{Bold} = better of the two, per group
  and metric.}
  \label{tab:cohort}
\end{table}

\section{Qualitative Galleries}
\label{sup:gallery}

This section extends Fig.~2 of the main paper. Every trace is read from the
committed evaluation caches; no panel re-runs a model. Predictions are drawn at
held-out positions only and joined to the two observed samples bracketing each
gap, so a method's curve inside a band is exactly what it contributed to that
window's in-gap RMSE.

\paragraph{How the windows are chosen.}
Panels are selected by a fixed rule rather than by eye. Within a
(mechanism, rate) cell we first discard windows whose gap count falls outside
$[0.5\times, 2\times]$ the cell's median gap count, which removes masks that are
structurally atypical for that mechanism; among the rest we take the window at
the median of CAIR's in-gap RMSE. The selection does not consult the
baselines. The resulting panels have CAIR-to-linear error ratios of
$0.85$--$1.01$, against $0.81$--$0.91$ for the mechanism-level aggregates in
Table~1 of the main paper.
Fig.~\ref{sup:fig:hard} is the one deliberate exception, sweeping the same cell
from its easiest to its hardest window.

\begin{figure*}[t]
  \centering
  \includegraphics[width=\textwidth]{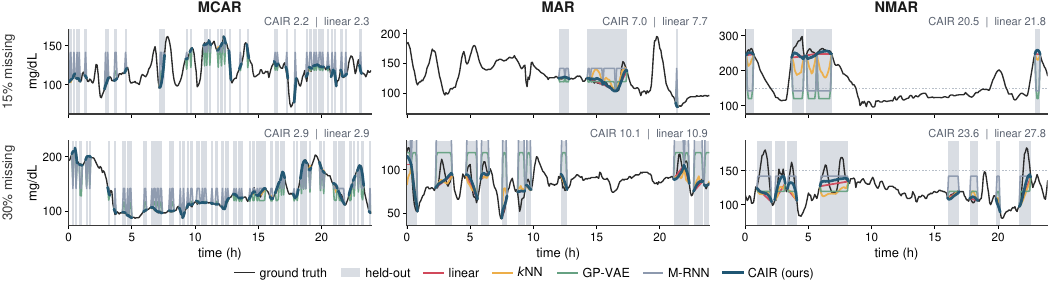}
  \caption{\textbf{AI-READI CGM, both missingness rates} (columns: mechanism;
  rows: rate). The main paper shows the $30\%$ row. Raising the rate from
  $15\%$ to $30\%$ lengthens and multiplies the blocks but does not change the
  ordering: CAIR\ and linear track the trace under MCAR, separate under MAR,
  and separate furthest under NMAR, where the dashed $70$/$150$\,mg/dL
  thresholds mark the excursions that trigger the deletion. GP-VAE and M-RNN
  revert to a near-constant fill under every mechanism, which is what
  puts them below every non-constant baseline in Table~1 of the main paper.}
  \label{sup:fig:rates}
\end{figure*}

\begin{figure*}[t]
  \centering
  \includegraphics[width=\textwidth]{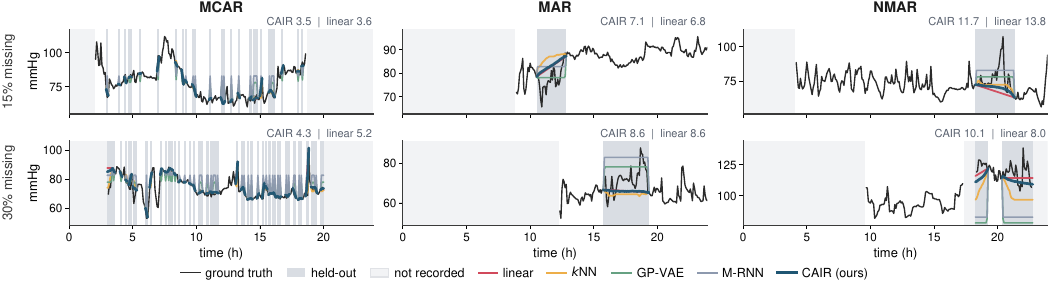}
  \caption{\textbf{MIMIC-III arterial pressure, same protocol.} Pale regions
  marked \emph{not recorded} are positions the ICU monitor never sampled
  ($23\%$ of ABP positions); they are neither observed nor scored, and no
  ground truth is drawn there. ABP is smoother and more locally linear than
  glucose, so the per-window gap between CAIR\ and linear is small and its
  sign varies from window to window (three of these six median-difficulty
  panels favour CAIR, two favour linear and one is a tie), while the
  six-rate aggregate in Table~6 of the main paper favours CAIR\ under all
  three mechanisms. The neural baselines fail here exactly as they do on CGM.}
  \label{sup:fig:mimic}
\end{figure*}

\begin{figure*}[t]
  \centering
  \includegraphics[width=\textwidth]{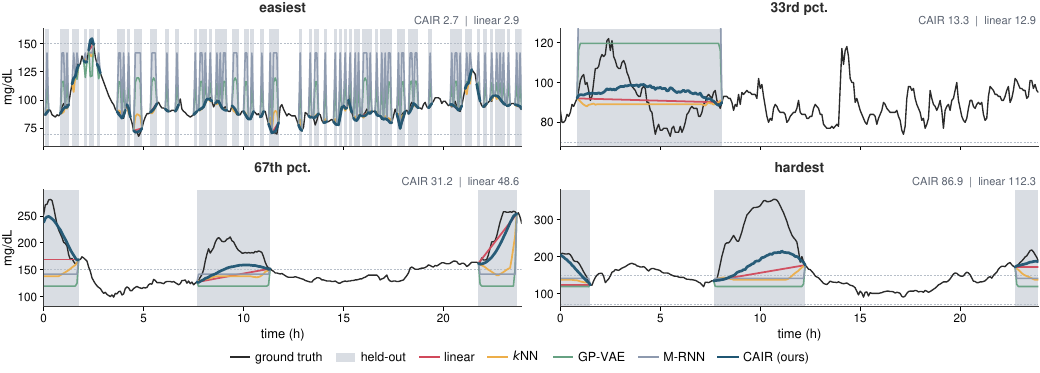}
  \caption{\textbf{Easiest to hardest, at fixed mechanism and rate}
  (AI-READI CGM, NMAR, $30\%$; panels at the $0$th, $33$rd, $67$th and $100$th
  percentile of CAIR's in-gap RMSE). These panels support the claim in
  Sec.~4.8 of the main paper that CAIR\ degrades gracefully. As the blocks
  lengthen and swallow whole excursions, no method recovers the excursion, but
  the failures differ in kind: GP-VAE and M-RNN sit at a constant fixed by the
  training distribution, linear draws the chord its endpoints imply, and
  CAIR\ bends part of the way toward the excursion and stops. In the hardest
  panel the recorded in-gap peak is $354$\,mg/dL; CAIR\ reaches $215$ and linear
  $176$. Under-shooting is the safe direction of error for a
  clinical burden metric, since an invented excursion would create a treatment
  signal that never occurred.}
  \label{sup:fig:hard}
\end{figure*}